\documentclass[journal]{IEEEtran}
\usepackage{cite}
\usepackage{amsmath,amssymb,amsfonts}
\usepackage{algorithmic}
\usepackage{graphicx}
\usepackage{textcomp}
\usepackage{threeparttable} %http://blog.modelworks.ch/tables-with-footnotes-in-latex/
\usepackage{footnote}
\usepackage{multirow} %https://blog.csdn.net/lishoubox/article/details/7330409
\usepackage{pifont}% http://ctan.org/pkg/pifont
\usepackage{times}
\usepackage{color, colortbl}  %the first to define new colors and the latter to actually color the table
\usepackage[table]{xcolor}
\usepackage{graphicx}
\usepackage{bbding}
\usepackage{soul}
\usepackage[utf8]{inputenc}
\usepackage{overpic}
\usepackage{booktabs}
\usepackage{diagbox}
\usepackage{etoolbox}
\usepackage{url}
\usepackage{rotating}
\apptocmd{\sloppy}{\hbadness 10000\relax}{}{}
\usepackage[multiple]{footmisc}
\definecolor{peach-orange}{rgb}{1,0.8,0.6}
\definecolor{light_blue}{rgb}{0.6,0.81,0.93}
\usepackage[colorlinks=true,pdftex]{hyperref} %\usepackage[hidelinks]{hyperref}  %
\newcommand{\figref}[1]{Fig.~\ref{#1}}%
\newcommand{\tabref}[1]{Table~\ref{#1}}%

\renewcommand{\eqref}[1]{Eq.~(\ref{#1})}

\newcommand{\tabincell}[1]{\begin{tabular}{@{}c@{}}#1\end{tabular}}   %replace c with #1 for alignment control

\newcommand{\CorrectionTextColor}{black}
\newcommand{\CorrectionTextColorSecond}{black}

\def\BibTeX{{\rm B\kern-.05em{\sc i\kern-.025em b}\kern-.08em
    T\kern-.1667em\lower.7ex\hbox{E}\kern-.125emX}}
\begin{document}
\title{ CASIA-Face-Africa: A Large-scale African Face Image Database }
\author{Jawad~Muhammad, Yunlong~Wang,  \IEEEmembership{Member, IEEE}, Caiyong~Wang, Kunbo~Zhang, \IEEEmembership{Member, IEEE}, and Zhenan~Sun, \IEEEmembership{Senior Member, IEEE}
\IEEEcompsocitemizethanks{
\IEEEcompsocthanksitem This work was supported in part by the National Natural Science Foundation of China under Grant U1836217, the National Key Research and Development Program of China under Grant 2017YFC0821602, 62006225 and the CAS-TWAS president's Fellowship for International Doctoral Students.  (\textit{Corresponding author: Zhenan~Sun.})
\IEEEcompsocthanksitem Jawad~Muhammad, Yunlong~Wang, Kunbo~Zhang and Zhenan~Sun are with the School of Artificial Intelligence, University of Chinese Academy of Sciences, Beijing 100049, China, and also with the Center for
Research on Intelligent Perception and Computing, National Laboratory of Pattern Recognition, Institute of Automation, Chinese Academy of Sciences, Beijing 100190, China (e-mail: jawad@cripac.ia.ac.cn; yunlong.wang@cripac.ia.ac.cn; kunbo.zhang@ia.ac.cn; znsun@nlpr.ia.ac.cn).
\IEEEcompsocthanksitem Caiyong Wang is with the School of Electrical and Information Engineering, Beijing University of Civil Engineering and Architecture,
Beijing 100044, China and also with Beijing Key Laboratory of Intelligent Processing for Building Big Data, Beijing 100044, China (e-mail: wangcaiyong@bucea.edu.cn).
\IEEEcompsocthanksitem \copyright 20XX IEEE.  Personal use of this material is permitted.  Permission from IEEE must be obtained for all other uses, in any current or future media, including reprinting/republishing this material for advertising or promotional purposes, creating new collective works, for resale or redistribution to servers or lists, or reuse of any copyrighted component of this work in other works.
}
}

\markboth{ }%
{Shell \MakeLowercase{\textit{et al.}}: Bare Demo of IEEEtran.cls for IEEE TIFS}

\maketitle

\begin{abstract}

Face recognition is a popular and well-studied area with wide applications in our society. However, racial bias had been proven to be inherent in most State Of The Art (SOTA) face recognition systems. Many investigative studies on face recognition algorithms have reported higher false positive rates of \textcolor{\CorrectionTextColor}{African subjects cohorts than the other cohorts}. Lack of large-scale African face image databases in public domain is one of the main restrictions in studying the racial bias problem of face recognition. To this end, we collect a face image database namely \emph{CASIA-Face-Africa} which contains 38,546 images of 1,183 African subjects. Multi-spectral cameras are utilized to capture the face images under various illumination settings. Demographic attributes and facial expressions of the subjects are also carefully recorded. For landmark detection, each face image in the database is manually labeled with 68 facial keypoints. A group of evaluation protocols are constructed according to different applications, tasks, partitions and scenarios. The performances of SOTA face recognition algorithms without re-training are reported as baselines. The proposed database along with its face landmark annotations, evaluation protocols and preliminary results form a good benchmark to study the essential aspects of face biometrics for African subjects, especially face image preprocessing, face feature analysis and matching, \textcolor{\CorrectionTextColor}{facial} expression recognition, \textcolor{\CorrectionTextColor}{sex/age} estimation, ethnic classification, face image generation, etc. The database can be downloaded from our website\footnote{\href{http://www.cripacsir.cn/dataset/}{http://www.cripacsir.cn/dataset/}}.

\end{abstract}


\section{Introduction}
\label{sec:introduction}
\IEEEPARstart{I}{n} recent years, there have been significant contributions to the research of face biometrics with many approaches proposed to solve the remaining and emerging problems. Solutions to such challenges including partial or occluded face detection\cite{Li18}, 3D facial landmark detection\cite{Zhang36}, face frontalization\cite{Cao2}, facial de-makeup\cite{Li17}, heterogeneous face recognition\cite{Liang19}, pose-invariant face recognition\cite{Li16}, kinship identification \cite{Robinson28}, liveness detection\cite{Rossler29}, facial aging\cite{Liu21} and age estimation\cite{Li15}, etc. Apart from the above developments, face image databases supporting these achievements have also been published and made publicly available to researchers. This has gradually become a trend in this deep learning era. Some of these popular face databases include LFW\cite{Huang12}, CASIA-Webface \cite{Yi35}, VGGFace2\cite{Cao3}, Ms-Celeb-1m\cite{Guo11}, MegaFace\cite{Kemelmacher13}, FRGC\cite{Phillips24}, CAS-PEAL\cite{Gao9}, MUCT\cite{Milborrow22}, etc.

However, researchers seldom concentrate on the racial bias problem despite its existence in most current face recognition systems \cite{Patrick2019Face}, \cite{Buolamwini1, Du6, Wang33}, \cite{Wang32}, \cite{Singer31}, \cite{Nagpal23}, \cite{cook2019demographic}, \cite{Howard2019The}, \cite{Krishnapriya2020Issues}, \cite{Drozdowski2020Demographic}. It is commonly known that people easily recognize faces from their own race, rather than faces from other races (\emph{aka.} other-race effect). Predictively, this phenomenon have been observed in the deployed face recognition systems \cite{Furl8}. \textcolor{\CorrectionTextColor}{Various ISO/IEC projects such as ISO/IEC 22116\cite{ISO201522116} have been dedicated to the study and documentation of this problem}. In a large evaluation study recently released by NIST \cite{Patrick2019Face}, it was reported that majority of the 106 face recognition algorithms have relatively higher false positive rates on subjects from African and Asian countries than those of other races. In addition, it also indicates that the algorithms developed in China usually attain much lower false positives on Asian subjects. Similar results have been previously investigated by other researchers. In \cite{Cavazos4}, the performance differences among the face recognition methods are considered as a function of the race that a face belongs to. In a study conducted by the American Civil Liberties Union (ACLU), the Amazon’s Recognition Tool misclassified photos of almost 40\% of the non-Caucasian U.S. congressmen as criminals \cite{Singer31}. In \cite{Buolamwini1}, some of the popular commercial facial classification systems were evaluated on the widely-used face datasets ( IJB-A, Adience and PPB) and it was found that the misclassification error of lighter skin subjects is much lower than that of the darker skin subjects. The recognition performance of lighter skin achieves only a maximum of 0.8\% error rate while the result on darker skin rises to a maximum error rate of 34.7\%. Besides, \cite{Wang33} declares that the face recognition rate on the African race is the lowest compared to the other race categories. In another \textcolor{\CorrectionTextColor}{study} in \cite{Wang32}, the authors experimentally \textcolor{\CorrectionTextColor}{verified} that subjects of black race and female cohort consistently show lower matching accuracy. In \cite{Phillips25}, the researchers combine five \textcolor{\CorrectionTextColor}{W}estern algorithms and five \textcolor{\CorrectionTextColor}{E}ast Asian algorithms to investigate the fusion results. Their results show that the western algorithm recognizes Caucasian faces more accurately than Asian faces and the Asian algorithm recognizes Asian faces more accurately than Caucasian faces, thereby affirming the ‘other-race effect’ characteristics.

It is evident from the existing studies that race plays an important role in  face recognition performance. However, despite Africans been projected to surpass 1.65 billion by the year 2030 (About 20\% of the projected world population) \cite{un2015population}, African race continually suffers the negative consequences from the racial bias problem. The NIST study report \cite{Patrick2019Face} suggests that algorithm developers should investigate the utility of more diverse, globally derived training data. Therefore, this paper proposes a large-scale African face image database. \textcolor{\CorrectionTextColor}{The face images were collected from African subjects using multi-spectral cameras in Nigeria, Africa}. We believe that it will serve as a comprehensive benchmark for tackling the problems of racial bias regarding the African cohorts. The contributions of this paper are summarized as follows:

\begin{enumerate}
  \item A large-scale face image database named \emph{CASIA-Face-Africa} captured from 1,183 African subjects using multi-spectral cameras is constructed and published for academic research. To the best of our knowledge, the proposed database is the first large-scale face database of African subjects available in public domain. Demographic attributes and facial expressions of the subjects in the \textcolor{\CorrectionTextColor}{database were} carefully recorded. It is expected to spur the research on the racial bias problem \textcolor{\CorrectionTextColor}{and enhance the performance of face recognition systems} on black race in real-world applications.
  \item Manual annotations of facial landmarks are provided to study the problems of facial keypoint detection, human-computer interaction and other related topics for African \textcolor{\CorrectionTextColor}{race}. A tailored software with GUI is developed for efficient labeling and editing of the facial keypoints. This GUI tool will be released together with the database.
  \item A group of evaluation protocols is constructed to split the data and objectively measure the performances of face recognition algorithms on the database. These protocols are specifically designed to cover possible applications, tasks, partitions and scenarios of the database, which will guarantee fair evaluations when comparing different face recognition algorithms.
  \item \textcolor{\CorrectionTextColor}{ The performance results of pre-trained models (adopted without any additional fine tuning or training on the database) of some representative face recognition methods on the \emph{CASIA-Face-Africa} are provided for comparison. The results demonstrated that }racial bias does exist in current SOTA face recognition methods and \textcolor{\CorrectionTextColor}{ there is need for bias mitigations among algorithms}.

\end{enumerate}

The remainder of this paper is organized as follows\textcolor{\CorrectionTextColor}{:} In section II, related work in the literature  will be discussed. Section III describes the proposed database in detail and the baseline evaluation performance on the database is provided in section IV. Section V concludes the paper and suggests future work.

\section{Related Work}
\label{sec::related_Work}
\subsection{Face Image Databases}
The face image databases publicly available can be broadly categorized into two groups: (1) constrained databases; and (2) unconstrained or in the wild databases. \textcolor{\CorrectionTextColor}{For constrained databases, the samples in the dataset are captured under controlled or prescribed imaging conditions and device configurations. In contrast, the images that constitute the unconstrained or in-the-wild databases are either downloaded directly from the Internet or scanned into digital copies from printed photographs, i.e, the process of image formation is arbitrary.  Recently, to address the demographic bias problems, some unconstrained datasets have been proposed that are subsets of the existing datasets such as: Gender-FERET\cite{Azzopardi2016Gender}, a subset of FERET\cite{Phillips26} with a balanced number of sex; DiveFace\cite{morales2020sensitivenets}, a subset of MegaFace\cite{Kemelmacher13} with a balanced number of sex and race;  DiF \cite{2019Diversity} and FairFace \cite{2019FairFace}, selected images from YFCC100M\cite{2017Learning} dataset with balanced number of age, sex and race subjects;   DemogPairs\cite{2019DemogPairs}, a subset of VGGFace2~\cite{Cao3} with a balanced number of sex and race; and RFW\cite{Wang33}, a subset of Ms-Celeb-1m~\cite{Guo11} with a balanced number of race; }

\textcolor{\CorrectionTextColor}{ For an effective demographic bias study, there is need for controlled imaging conditions across subjects of which factors such as capturing device, environmental conditions, and image resolution that can affect subjects intra- and inter- class distances are kept relatively constant for all the participating subjects. This will embolden other factors such as subject demographic attributes thereby making it easier to effectively study the effect of these attributes on the system performance outcome. From this viewpoint, the database presented in this paper belongs to the category of constrained database.}

In \tabref{tab:databases}, the recently published face image datasets are presented with vital details such as their racial distribution. Firstly, it can be observed that many of the widely-used databases do not include the racial distribution of the subjects. This is mainly due to the fact that the faces were downloaded from the Internet without racial information which will be very difficult to deduce. Among these databases, MORPH~\cite{Ricanek27} has the highest number of black subjects and images. However, this database is not readily available to researchers (payment required). Additionally, it does not account for a balanced male-female ratio, subjects with expression variations, and multi-source cameras to capture face images across the spectrum. The underlying reasons are that the images in MORPH are generated by digitization of the photographs through scanning. On the other hand, \textcolor{\CorrectionTextColor}{the majority of existing constrained databases have a }very limited number of black subjects and corresponding images.
\begin{table*}[!t]
  \begin{center}
  \renewcommand{\arraystretch}{1.1}
  \setlength\tabcolsep{3pt}
  \caption{Statistics of some popular face recognition databases and their racial distribution.}\label{tab:databases}%
  \begin{tabular}{p{0.15\linewidth}p{0.30\linewidth}p{0.08\linewidth}p{0.08\linewidth}p{0.03\linewidth}p{0.03\linewidth}p{0.12\linewidth}p{0.03\linewidth} }
    \toprule
    Datbase &   Notes &  Subject  &  Images & \tabincell{$E$} & \tabincell{$H$}  &  Race Distribution & \tabincell{$L$} \\
      \hline
      \multicolumn{8}{c}{Unconstrained or In the wild databases } \\
      \hline
    MORPH~\cite{Ricanek27} & Images are generated from scanning of photographs, and payment is required.  &  13,618 & 55,134 &  - & N  & A: 10, 519 (77\%) W: 2,608; AS: 38; H: 436  & -\\
       \hline
    PPB~\cite{Buolamwini1} & Images are from parliamentarians posted on government websites.  it can only be used for \textcolor{\CorrectionTextColor}{sex} classification.   &  1270   & 1270 &  - & N  & A: 661 (52\%) W: 609 & -\\
       \hline
          \tabincell{IJB-A~\cite{Klare14}} & Images are collected from the Internet.   &  500 & 5,712 &  - & N  & A: 102 (20.4\%) W: 398 & 3\\
       \hline
          \tabincell{AdienceDb~\cite{Eidinger7}} & Images are downloaded from Flickr.com albums. The landmark points are predicted automatically, but age, \textcolor{\CorrectionTextColor}{sex} and identity are manually labeled   &  2,284 & 26,580 &  - & N  & A: 302  (13.8\%) W:  1892  & 3\\
       \hline
        \tabincell{VGGFace2~\cite{Cao3}} & Images are downloaded through Google Image Search. The images have large variations in pose, age,  illumination, ethnicity and profession.   &  9,131 & 3,310,000 &  - & N  & - & 3\\
       \hline
       \tabincell{CASIA-Webface~\cite{Yi35}} & Images are downloaded from the IMDb website containing rich information of celebrities.  Celebrities born ranging from the year 1940 to 2014 are selected without considering their nationality or race.   &  10,575 & 494,414 & - & N  & - & -\\
       \hline
       \tabincell{ Ms-Celeb-1m~\cite{Guo11} } & it consists of 10 million face images and is harvested from the Internet. Majority of the identities are American and British actors.   & 100,000 & 8,200,000 & - & N  & - & -\\
       \hline
       \tabincell{ MegaFace\cite{Kemelmacher13} } & Images are downloaded from Yahoo Flickr website.   & 672,057 & 4,700,000 & - & N  & - & -\\
       \hline
       \tabincell{ WIDER FACE\cite{yang2016wider} } & Images are collected through Google and Bing searches.   & - & 32,203 & - & N  & - & -\\
       \hline
       \tabincell{ AgeDB-30\cite{moschoglou2017agedb} } & It is manually collected from the Internet without using web crawlers. Explicit image attributes are stated.   & 568 &  16,488 & - & N  & - & -\\
       \hline
       \tabincell{ CFP-FP\cite{sengupta2016frontal} } & It is made up of  images downloaded from the internet using keywords, most of the faces are belonging to politicians, athletes and entertainers.   & 500 &  20,000 & - & N  & - & -\\
        \hline
       \tabincell{ \textcolor{\CorrectionTextColor}{MEDS II\cite{Founds2010NIST} } } &  \textcolor{\CorrectionTextColor}{A test corpus organized from an extract of submission files of deceased persons containing biographic and biometric data recorded during an encounter of an individual.}   & \textcolor{\CorrectionTextColor}{518} &  \textcolor{\CorrectionTextColor}{1309} & - & \textcolor{\CorrectionTextColor}{N}  & \textcolor{\CorrectionTextColor}{A: 195(37.6\%); W:260; AS: 8; AS-S: 10;}  & \textcolor{\CorrectionTextColor}{68} \\
       \hline
       \multicolumn{8}{c}{Constrained databases} \\
      \hline
      \tabincell{MUCT~\cite{Milborrow22}} & The images are captured at University of Cape Town south Africa with many black subjects and other races. However, the race information is not publicly available.   &  276 & 3755 &  - & N  & - & 76 \\
       \hline
      \tabincell{M2FPA~\cite{Li16}} & It is suitable for facial pose analysis. Images are captured using multiple cameras positioned at several angles so that multiple poses are captured.   &  229 & 397,544 &  2 & Y  & A: 0 (0\%); AS: 229 & - \\
       \hline
      \tabincell{CAS-PEAL-R1~\cite{Gao9}} & It is a large dataset captured using an array of cameras. Multiple poses, light settings  and expressions are captured simultaneously for each subject.   &  1040 & 99,594 &  5 & Y  & A: 0 (0\%); AS: 1040 & - \\
       \hline
      \tabincell{FRGC~\cite{Phillips24}}-validation set & It comprises of 3D images and still images captured under controlled and  uncontrolled conditions.   &  466 & - &  2 & Y  & A: 47 (10\%); W: 316;  AS:103 & -\\
       \hline
      \tabincell{ Multi-pie\cite{gross2010multi} } & Images are captured with high-resolution cameras over 4 recording sessions with varying poses and illumination.  & 337 & 755,370 & 6 & Y  & A:10(3\%); W:202; AS:118; O:7 & -\\
       \hline
      \tabincell{ColorFERET~\cite{Phillips26}}-validation set & It is captured in a semi-controlled environment over multiple sessions  distributed by NIST.   &  994 & 11, 338 &  2 & N  & A: 78 (7.8\%); W: 618; AS:   171; AS-ME: 53; AS-S: 1; H: 57;N-A: 2; P-I: 10; Others: 4;  & 4 \\
      \hline
      \tabincell{ CASIA-Face-Africa } & Our proposed Dataset.  & 1183 & 38,546 & 7 & Y  & A:1183(100\%)  & 68\\
       \bottomrule
   \end{tabular}
   \begin{tablenotes}
    E: No. of Expressions; 	H: Heterogeneous images uniform across all subjects; 	L: No. of labeled points; A: 'Black-or-African-American'; 	W: 'White'; 	AS: 'Asian'; 	AS-ME: 'Asian-Middle-Eastern'; 	AS-S: 'Asian-Southern'; 	H:  'Hispanic'; 	N-A: 'Native-American'; 	P-I: 'Pacific-Islander' O: 'Others'
    These race labels has been loosely adopted from \cite{Phillips26}, consistent with their definitions.
    %\item[a] for the abstraction reaction, $\fam0 Mu+HX \rightarrow MuH+X$.
    %\item[b] 1 degree${} = \pi/180$ radians.
   \end{tablenotes}
   \end{center}\vspace{-10pt}
\end{table*}

\subsection{Face Detection}
Face detection is indispensable in the pre-processing of face images, which also affects the performance of all subsequent processes. As a \textcolor{\CorrectionTextColor}{subfield}, face detection \textcolor{\CorrectionTextColor}{naturally} keeps evolving with object detection. Numerous approaches have been proposed for this task, starting from the popular pioneering work of Viola and Jones \cite{viola2001rapid} to the recently proposed deep-learning based methods. In analogy with other object detection tasks, the deep-learning methods for face detection can be  categorized into: (1) Single-stage networks; and (2) Two-stage networks.

Typically, the two-stage networks involve \textcolor{\CorrectionTextColor}{the} adoption of an additional sub-network or external path that is not part of the main branch. They usually perform secondary tasks such as region proposals, filtration of invalid regions, etc. \textcolor{\CorrectionTextColor}{A}pproaches like Fast RCNN \cite{girshick2015fast}, Faster RCNN \cite{ren2015faster},  MTCNN \cite{zhang2016joint}, fall into this category. In Fast RCNN, region proposals are done via selective search \cite{uijlings2013selective} and classification on feature maps. Instead of selective search, a region proposal network in Faster RCNN is independently trained. The features of each region are extracted using ROI pooling technique. In MTCNN \cite{zhang2016joint}, four cascaded networks were adopted to classify and detect the faces with their landmark points based on repetitive calibration of the generated bounding box regions. MTCNN is effective but relatively slow due to the extra computations from the selection of multiple regions, classification stages and image rescaling.

In contrast, the single-stage networks are usually trained in \textcolor{\CorrectionTextColor}{an} end-to-end manner with no additional independently-trained module. Recently, several one-stage approaches such as S\textsuperscript{3}FD \cite{zhang2017s3fd}, DSFD \cite{li2019dsfd} and YOLOFace \cite{chen2020yolo} have been proposed to advance the \textcolor{\CorrectionTextColor}{research on }face detection. The problem of the region proposal in \textcolor{\CorrectionTextColor}{a} one-stage network is solved through tiling of the CNN feature maps. Moreover, the bounding boxes on multiple layers of the CNN feature hierarchy are taken into account to predict the localization of faces at various scales. These single-stage methods are generally fast due to their end-to-end inference. However, they are less accurate than the two-stage networks. In this paper, the methods MTCNN, S\textsuperscript{3}FD, DSFD and Yoloface are selected as baselines to evaluate the performances of face detection on the proposed African database.

\subsection{Facial Representation}
With the development of deep-learning frameworks, \textcolor{\CorrectionTextColor}{profound success have been} witnessed in the \textcolor{\CorrectionTextColor}{aspect} of facial representation and feature analysis. Some of the \textcolor{\CorrectionTextColor}{effective} methods prior to the deep-learning era include Eigenface \cite{turk38}, Fisherface (LDA) \cite{belhumeur37}, Gabor+LDA \cite{liu39}, etc. But, these \textcolor{\CorrectionTextColor}{techniques} are usually handcrafted and require manual tuning of the parameters. As such, they are generally less effective than the deep-learning methods in automatic facial representation.
Typically, the deep-learning methods adopt Convolutional Neural Networks (CNN) to extract a fixed-length feature vector as facial representation from cropped and aligned face regions. In other words, face images are embedded into the same feature space where each identity is formulated as one cluster. In this paper, \textcolor{\CorrectionTextColor}{some of} the embedding-based models are employed such as Facenet \cite{Schroff30}, SphereFace \cite{Liu20}, LightCNN \cite{Wu34} and Arcface \cite{Deng5} to evaluate the generalization ability of recent representative methods on the proposed African dataset. In Facenet, face images are projected into an embedding of compact Euclidean space where \textcolor{\CorrectionTextColor}{the distance between two embeddings} corresponds to their face similarity \cite{Schroff30}. The parameters of the network were learnt directly from the training face images and optimized by Euclidean margin based triplet loss function. In SphereFace, instead of the Euclidean margin, an angular margin-based loss function is employed for training \cite{Liu20}. The angular margin function is a modification of the Softmax function through manipulation of its decision boundaries. In ArcFace, an additive angular margin is imposed instead, to enhance the extracted embedding features \cite{Deng5}. The extracted embedding has a clear geometric interpretation because of its correspondence to the geostatic distance on the hyperspace. In LightCNN, the embedding is \textcolor{\CorrectionTextColor}{generated} through Max-Feature-Map (MFM) operations which produce a compact representation of the input face image \cite{Wu34}.

\section{The Database}
\label{sec::database}
\subsection{Image Acquisition}
The face images of the proposed database were captured at both indoor and outdoor environments of collages and neighborhoods in Nigeria, Africa. \textcolor{\CorrectionTextColor}{The capturing locations are: Dabai city in Katsina state; Hotoro in Kano state; Birget in Kano state; Gandun Albasa in Kano state; Sabon Gari in Kano state; and Kano State School of Technology. These locations were strategically selected as they are known to have diverse population of local ethnicities.} About 1,200 volunteers participated in the capturing activity which was conducted during daytime and night. \textcolor{\CorrectionTextColor}{ The subjects voluntarily accept to participate and signed a consent form allowing their facial data to be used exclusively for academic research purposes.}
The images of each subject were captured concurrently using \textcolor{\CorrectionTextColor}{multi-spectral cameras of two visible wavelength (VW) cameras and one near-infrared (NIR) camera}. The NIR camera and one of the VW cameras \textcolor{\CorrectionTextColor}{were} bundled together into one device \textcolor{\CorrectionTextColor}{which makes} the number of camera modules \textcolor{\CorrectionTextColor}{ to be} two instead of three.

\begin{figure}[!tbp]
  \centering
  \begin{overpic}[width=0.8\linewidth]{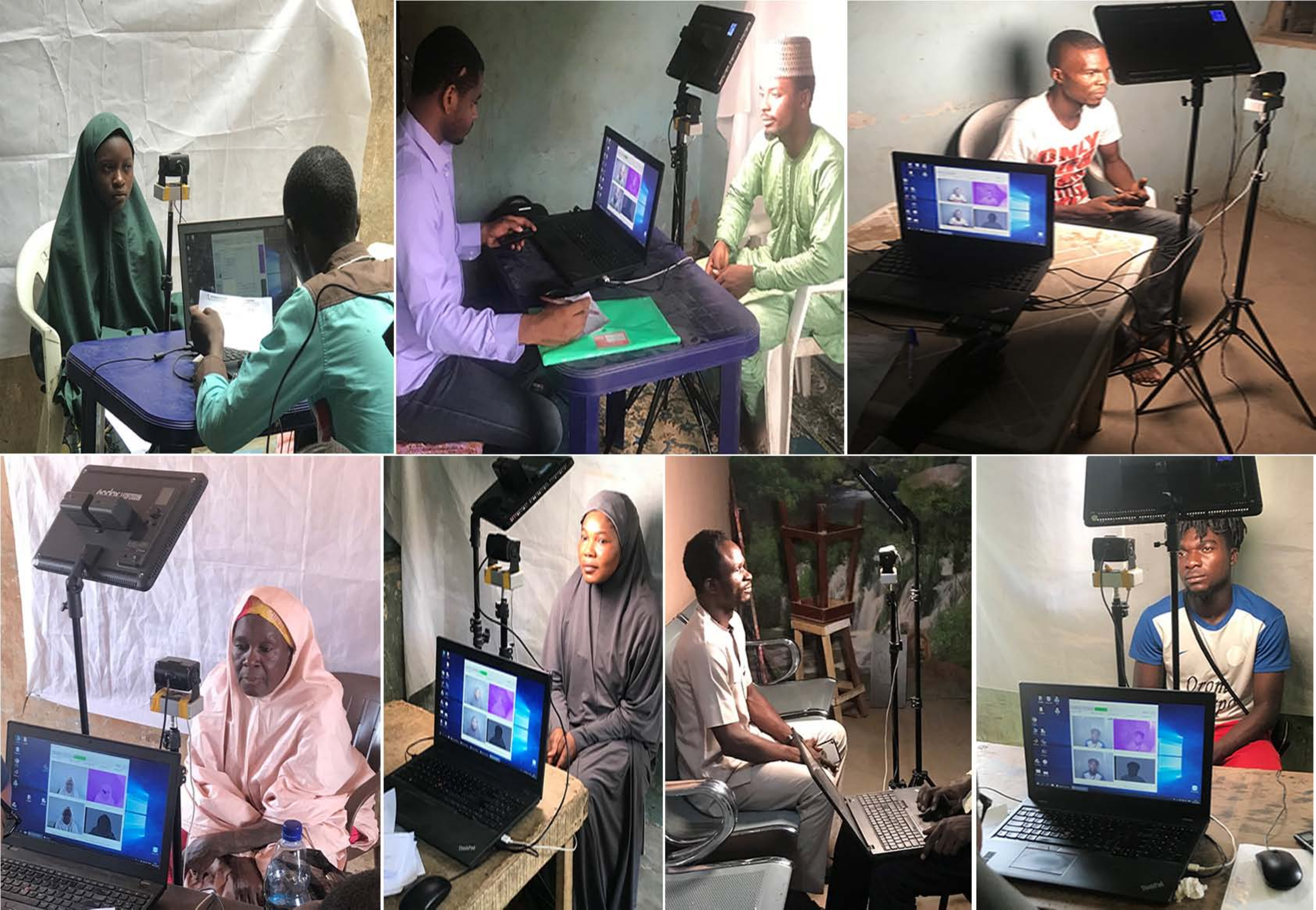}
  \end{overpic}\vspace{-3pt}
  \caption{The camera arrangement and set-up of capturing scenes.}
 \label{fig:camera_setup}\vspace{-10pt}
\end{figure}

The process of the image collection was completed in interval sessions over a period of three months. Some of the subjects have their face images captured in multiple sessions while the majority of the subjects have their face images captured in a single session. \textcolor{\CorrectionTextColor}{Specifically, only 17 subjects across the three camera category had their images captured in two sessions while all others have their images captured in one session.} The camera arrangement and set-up of capturing scenes are configured to be \textcolor{\CorrectionTextColor}{the} same in all the sessions as shown in \figref{fig:camera_setup}.

 For each subject, 3 to 10 still face images were captured by each camera at a duration of 1 to 3 seconds. Some of the subjects were guided to make 7 prescribed facial expressions i.e. \emph{Neutral}, \emph{Angry}, \emph{Sad}, \emph{Happy}, \emph{Surprise}, \emph{Fear} and \emph{Disgust} with the pictures of their faces captured simultaneously. A large portion of subjects also cooperate with an external illumination light source for additional image capture. Besides, some subjects are willing to wear face \textcolor{\CorrectionTextColor}{accessories} such as eye glasses and hat which increases the diversity of face images. Specifically, most of the African female subjects have their ears hidden behind some form of head covering (scarf). This is a traditional and local style of dressing for ages but such images are relatively scarce on the Internet.

\subsection{Database Organization}
\begin{figure}
 \centering
  \begin{overpic}[width=0.8\linewidth]{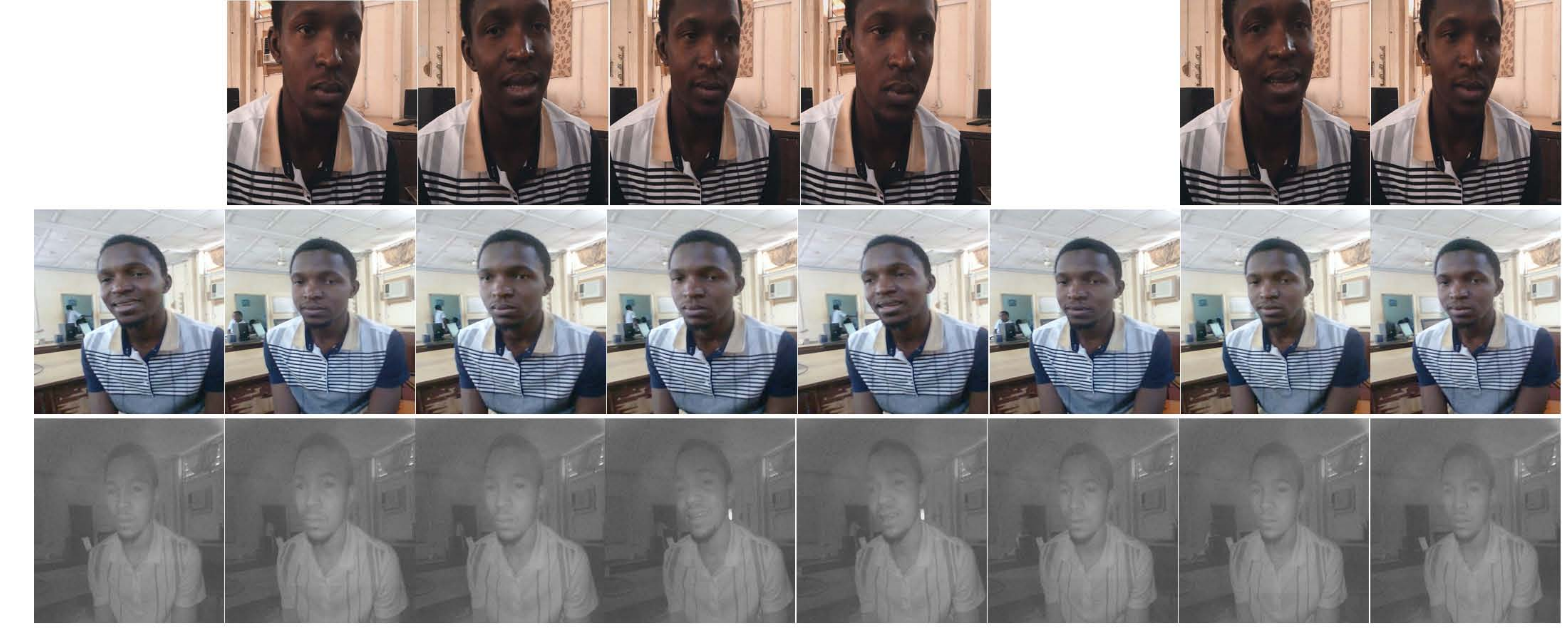}
  %\put (-1,5) {(a)}
  %\put (-1,19) {(b)}
  %\put (-1,32) {(c)}
  \end{overpic}
  \caption{A complete set of face images captured from one subject in a single session. The rows from top to down are images of VW Camera 1 (top), VW Camera 2 (middle), NIR camera (bottom).}
  \label{fig:sampleImages}

\end{figure}

\begin{figure}
\centering
  \begin{overpic}[width=0.8\linewidth]{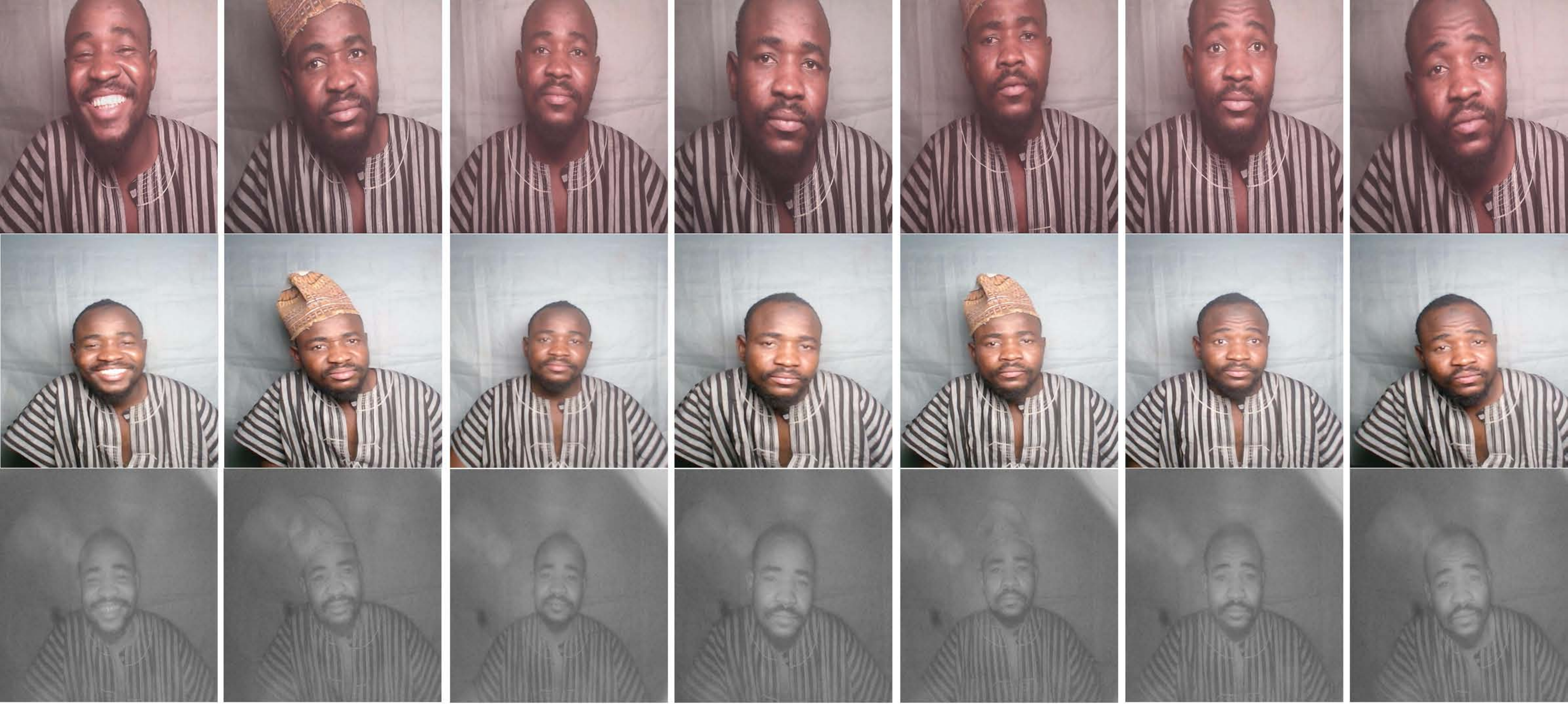}
  %\put (3,-1)  {Happy}
  %\put (15,-1) {Disgust}
  %\put (30,-1) {Neutral}
  %\put (44,-1) {Sad}
  %\put (59,-1) {Angry}
  %\put (74,-1) {Fear}
  %\put (86,-1) {Surprise}
  \end{overpic}
  \caption{Face images of 7 prescribed facial expressions from a single subject. The rows from top to down are VW camera 1, VW camera 2 and NIR camera respectively. The columns from left to right are the facial expressions in order, i.e. Happy, Disgust, Neutral, Sad, Angry, Fear and Surprise. }
  \label{fig:sampleImages_expresion}
  \vspace{-3pt}
\end{figure}

\begin{figure}
\centering
  \begin{overpic}[width=0.8\linewidth]{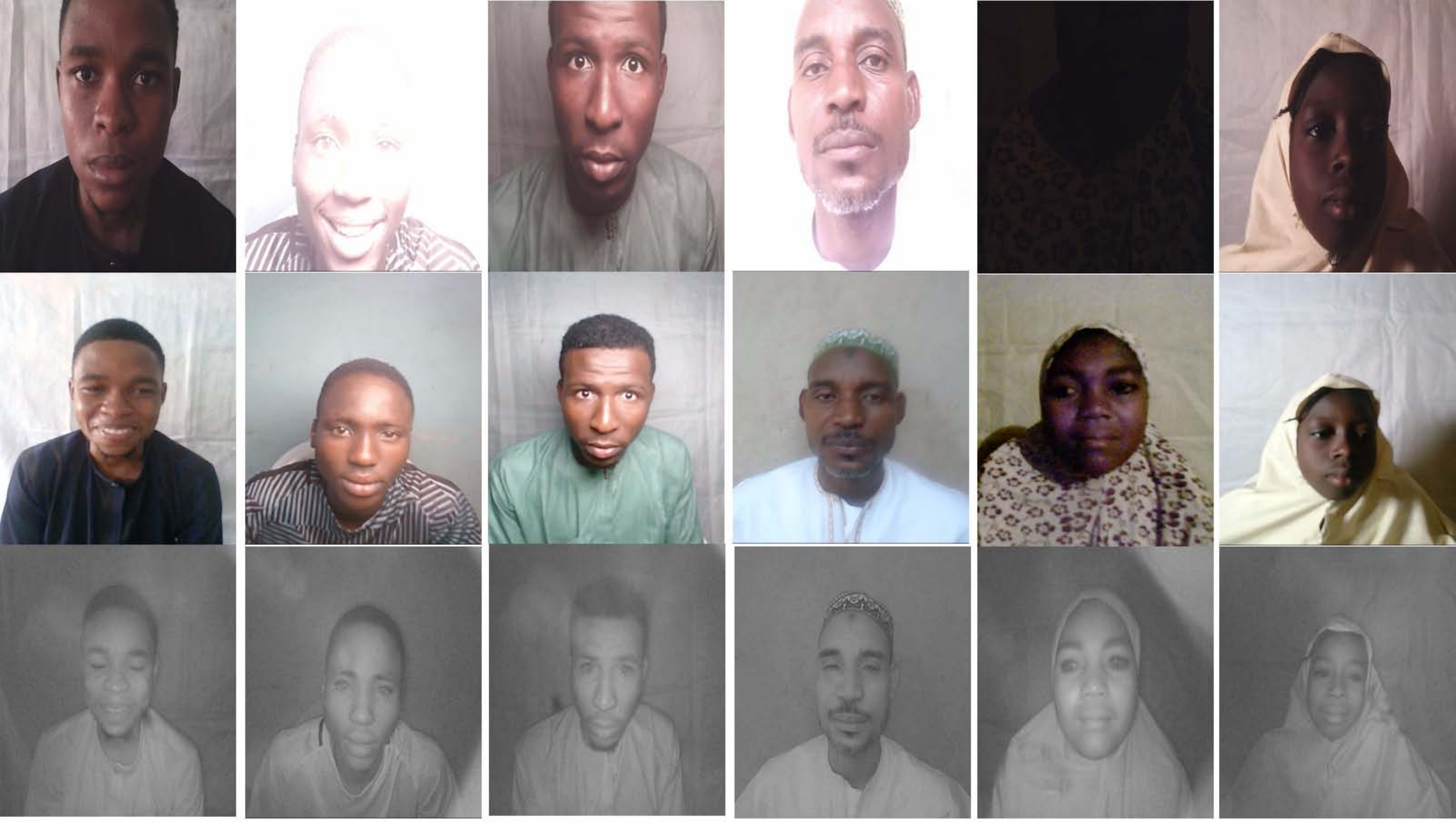}
  %\put (7,-1) {(a)}
  %\put (24,-1) {(b)}
  %\put (39,-1) {(c)}
  %\put (55,-1) {(d)}
  %\put (71,-1) {(e)}
  %\put (89,-1) {(f)}
  \end{overpic}
  \caption{Examples of the various capturing conditions. The rows from top to down are VW camera 1, VW camera 2 and NIR camera respectively. The columns from left to right are corresponding conditions, i.e. indoor day light, indoor day light with illumination, indoor night with illumination,  outdoor day light, outdoor night, and outdoor night with illumination.}
  \label{fig:illumination_conditions}
  \vspace{-3pt}
\end{figure}

The proposed database is composed of African face images, which are firstly selected from the captured raw data. Some face images of poor quality \textcolor{\CorrectionTextColor}{were} discarded due to imaging problems such as camera motion, over or underexposure, etc. Thereafter, the remaining images \textcolor{\CorrectionTextColor}{were} cropped to make sure that a face region is roughly centered in the image. A complete set of face images captured from one subject in a single session are depicted in \figref{fig:sampleImages}. Additionally, face images of 7 prescribed facial expressions from a single subject are shown in \figref{fig:sampleImages_expresion}. Examples of \textcolor{\CorrectionTextColor}{the various capturing conditions such as: daytime and night, indoor and outdoor, with or without extra illumination are shown} in \figref{fig:illumination_conditions}.


\begin{figure*}
    \begin{center}
    \includegraphics[width=0.9\linewidth]{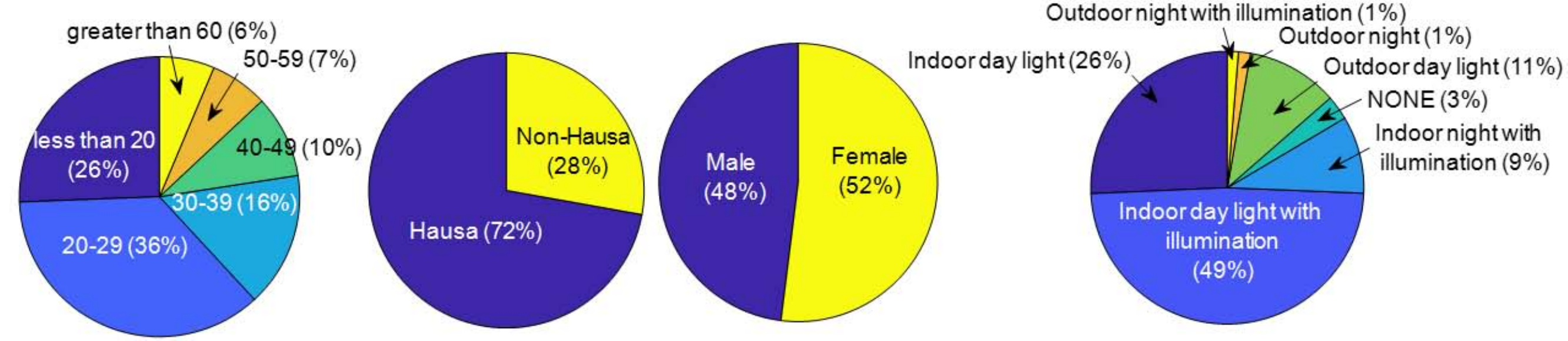}
    \end{center}
    \caption{Graphical statistics of the database. (a) Age, \textcolor{\CorrectionTextColor}{(b) Ethnicity, (c)Sex,} (d) Capturing Conditions.}
    \label{fig:database_summary}
\end{figure*}

The organized database comprises a total of  \textbf{38,546} images from  \textbf{ 1,183 } subjects after the selection. \textcolor{\CorrectionTextColor}{In particular, 12,063 images were captured by VW camera 1 at the resolution of $1332 \times 1080$. 13,232 images by VW camera 2 at the resolution of $787 \times 962$. And 13,251 images by NIR camera at the resolution of $983 \times 877$. \figref{fig:database_summary} shows the graphical statistical summary of the database.} It can be observed that:
\begin{itemize}
  \item The age distribution of the participating subjects as depicted in \figref{fig:database_summary} (a), indicates that the significant portion of the dataset comprises of the age groups 30$\sim$39, 21$\sim$29 and less than 20. This is considerate of their dominance on the workforce in most organizations.
  \item The Male-Female ratio of the dataset is 48\% - 52\%, \textcolor{\CorrectionTextColor}{which is almost sex-balanced} as depicted in \figref{fig:database_summary} (b). This \textcolor{\CorrectionTextColor}{can be useful for sex-orientated tasks}.
  \item The ethnicity distribution between Hausa ethnic group and Non-Hausa ethnic group is shown in \figref{fig:database_summary} (c). There are multiple ethnic groups in the database, which makes it suitable for \textcolor{\CorrectionTextColor}{a} study on ethnic attributes in Africa. Note that the number of the face images from Hausa ethnic group is much bigger than other ethnic groups. So it is appropriate to categorize Hausa as one cluster and all other ethnic groups as Non-Hausa. \textcolor{\CorrectionTextColor}{We believe having this large number of ethnic Hausa in the dataset also helps the dataset diversity as they are one of the most geographically dispersed ethnic groups in Africa, spanning multiple countries like Sudan, Chad, Binin, Ivory Coast, etc.}
  \item The summary of the capturing conditions is presented in \figref{fig:database_summary} (d) which shows that almost half of the images in the database are captured in an indoor setting with an external light illumination.

%   Specifically, as shown in \figref{fig:database_summary} (e), indoor images constitutes of almost 86\% with majority ( 88\% ) captured during the day as shown in \figref{fig:database_summary} (f). The use of illumination light source was in 61\% of the images as shown in \figref{fig:database_summary} (g).
\end{itemize}

\subsection{Database Labeling}
\begin{figure}[!tbp]
  \centering
  \begin{overpic}[width=0.9\linewidth]{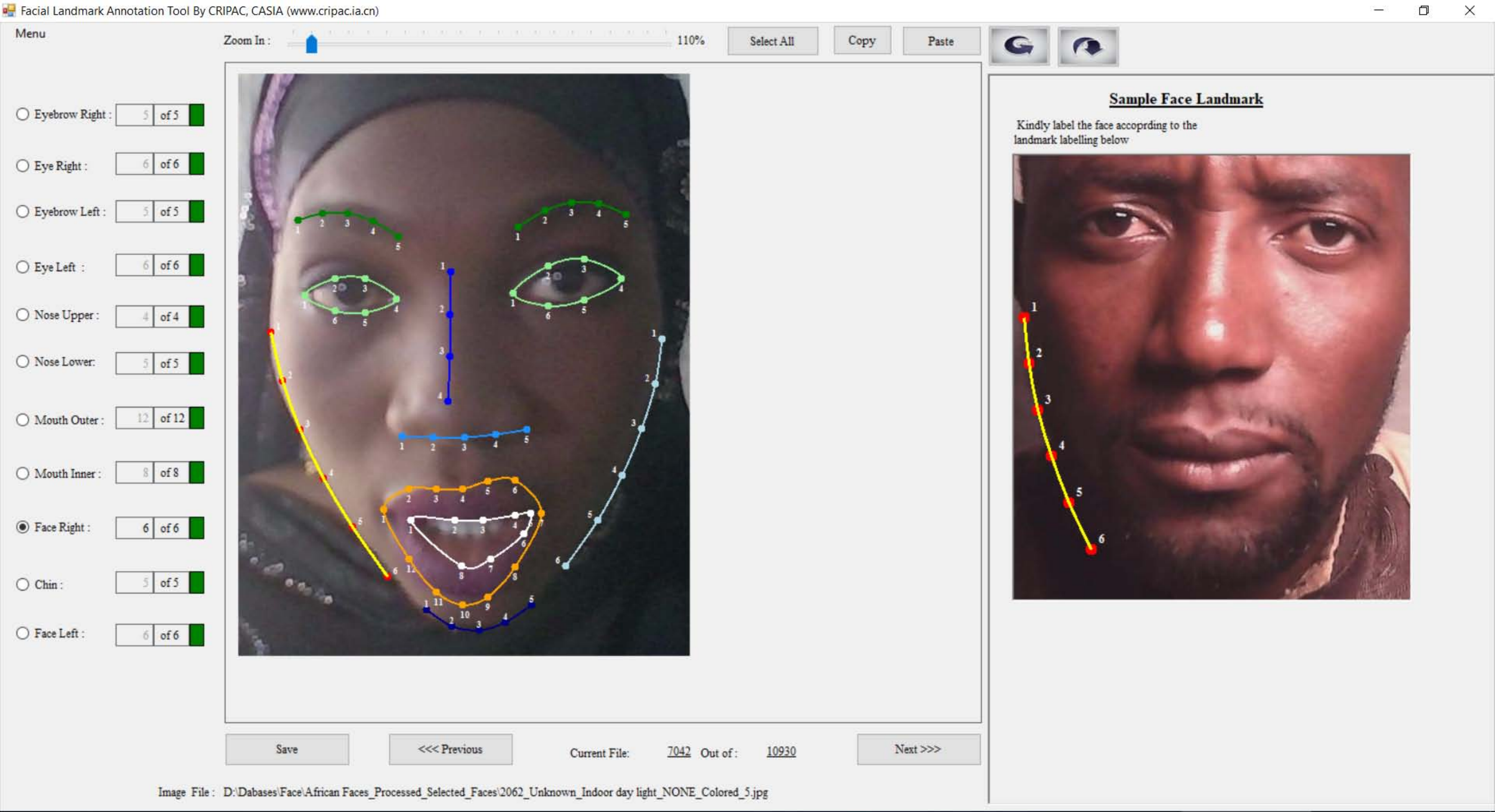}
  \end{overpic}\vspace{4pt}
  \caption{ Screenshot of the GUI tool developed for landmark annotation. }
 \label{fig:GUI}\vspace{-3pt}
\end{figure}

 \textcolor{\CorrectionTextColor}{Aside from } the provided identity and demographic attributes of each face image, typical 68 landmark points are also manually annotated \textcolor{\CorrectionTextColor}{which is based on the mark-up labeling in Multi-PIE \cite{Gross10}}. To accomplish this task, a \textcolor{\CorrectionTextColor}{semi-automated GUI tool was} developed which will be released with the dataset. A screenshot of the GUI is shown in \figref{fig:GUI}. The GUI tool can  certify the requirements of precision, swiftness, flexibility and user friendliness for face landmark annotation, specifically tailored for African face labeling. Some critical features of the developed GUI include:
\begin{itemize}
    \item A series of pre-defined rules regarding constraints and interrelations of facial landmarks are built-in so that each labeling point must certify these rules before it can be added. With these automatic inspections, labeling errors caused by careless operations will be greatly reduced.
    \item  \textcolor{\CorrectionTextColor}{Apart from} the built-in rules of facial landmarks, the GUI tool is able to fit the keypoints of the same region into a smooth curve and adjust these points in equal-interval positions.
    \item Grouping of points based on facial priors makes labeling of face sections (such as eye or mouth) faster and easier to manage.
    \item The GUI tool possesses superior ability to copy and paste the existing landmark points from one face image to another. Also, selected points can be freely rotated so as to accommodate various facial poses.
    \item Re-scaling or so-called zooming of the face image is well supported, regardless of its resolution and the keypoints already annotated.
    \item The landmark points already labeled can be edited (alteration or deletion) again as the user requires.
\end{itemize}
With the GUI, 68 landmark points of all the face images in the dataset are manually labeled and will be supplied for potential research.

\subsection{Proposed Protocols}
In order to facilitate the research on \emph{CASIA-Face-Africa}, a group of protocols have been carefully designed to provide seamless adaptation of the database. The principles behind the proposed protocols are based on the following aspects: (1) Application; (2) Task; (3) Partition and (4) Scenario. The naming convention of a protocol is the canonical combinations of these components as:
\emph{Application- Task - Partition  - Scenario}.

\subsubsection{Application}
A face image database may be used for various applications, e.g. verification, identification, facial expression recognition, \textcolor{\CorrectionTextColor}{age/sex} estimation, ethnic classification, etc. In the proposed protocols, the presumed applications of the database based on the available attributes of the subjects are:
%\begin{description}[\IEEEsetlabelwidth{Express}]
\begin{itemize}
\item \emph{Identity:} Person identification is the main application of the face database. The potential topics include but not limited to subject identification, verification, face image preprocessing, face feature analysis and matching, face liveness detection, face image generation, etc.
 \item \emph{Expression:} Affective computing is a bottleneck of machine intelligence now. It will be valuable to have a public African face database containing seven types of expressions namely \emph{Neutral, Angry, Sad, Happy, Surprise, Fear, and Disgust}. 79 subjects were captured with these facial expressions in the database.
\item \textcolor{\CorrectionTextColor}{\emph{Sex:} Sex} classification of African subjects based on face images is an important investigation topic. The database contains \textcolor{\CorrectionTextColor}{approximately }equal number of male and female subjects \textcolor{\CorrectionTextColor}{which make it suitable for sex classification study}.
\item \emph{Age:} Age estimation has become a research hotspot in recent years. The proposed database contains 63 distinct subject ages, but because of limited samples of some age groups, the database is coarsely grouped into 48 age categories.
\item \emph{Ethnicity:} Ethnicity classification is a featured topic as there are more than 50 ethnic groups in the database which are all native African races. However, due to the small number of the subjects in some of the ethnicity groups, only 12 groups with relatively large number of subjects are considered as separate groups and the remaining groups \textcolor{\CorrectionTextColor}{are} categorized as others, making up 13 ethnic groups in total. \textcolor{\CorrectionTextColor}{Only some subjects were} selected for this task, and thus the affiliated ethnic groups of these subjects are relatively balanced.
\end{itemize}

\subsubsection{Task}
The four basic tasks herein are as follows:
\begin{itemize}
  \item \emph{T:} The major part of face image database is usually reserved for algorithmic \emph{Training}. \emph{T} is the default symbol for all protocols of the training partition. Note that only this task is attributed to the training partition while the other three tasks are associated \textcolor{\CorrectionTextColor}{with} the testing partition
  \item \emph{\textcolor{\CorrectionTextColor}{I:} } \textcolor{\CorrectionTextColor}{This is the \emph{close set identification}. Each image in a query set $Q$ will be matched against the images in the target sets $T$. The computed scores will be ranked, and then the final identification result is inferred. $T$ comprises only of a single image per subject and the subjects in  $Q$ must be found in $T$. The single image was randomly selected from the image set of the subject. By analogy, the target-query setting here is equivalent to the genuine gallery-probe setting.}
  \item \emph{\textcolor{\CorrectionTextColorSecond}{O:} } \textcolor{\CorrectionTextColorSecond}{This is the \emph{open set identification}. This is similar to \emph{close set identification} except that some subjects in $Q$ are not necessarily contained in the $T$ set. As such, the $Q$ set is equivalent to the combination of genuine and imposter probes of the open set identification problem.}
  \item \emph{V:} This is the \emph{Verification}. Each image in a query set $Q$ is matched with each image in the target set $T$. The corresponding similarity scores constitute the verification results. This task can be performed by using all possible pairs of $T$ versus $Q$ combinations.
  \item \emph{C:}  This is the \emph{Classification}. Unlike the above two tasks \textcolor{\CorrectionTextColor}{of} recognizing subjects based on their unique attributes like identity, this task involves categorization of images based on the demographic or behavioral attributes of the subjects such as age, \textcolor{\CorrectionTextColor}{sex}, ethnicity and expression. An image in the test set $T$ is classified or clustered into the predefined classes by a classifier.
\end{itemize}

\subsubsection{Partition}
Consistent with the potential tasks, the partition of the database is mainly in two modes:
\begin{itemize}
  \item \emph{All:} This involves the adoption of all subjects and their images in the database for either training or testing but not both. Therefore, no portion of the database will be reserved for any other tasks. If it is used for only training, the database is served as supplementary data to enhance the African biometric performance of a specific method. Otherwise, if it is used for testing, the whole database will be treated as a testing partition.
  \item \emph{Split:} This involves the splitting of the database into training and testing partitions. In each imaging category, about  40\% of the subjects have been selected into the training partition and \textcolor{\CorrectionTextColor}{the remaining 60\% of the subjects selected into testing partition. There is no overlap of subjects between the two partitions} . Regarding the details of the protocols, the specific list of the selected subjects as well as their face images and configuration files for each partition will be released together with the dataset.

\end{itemize}

\subsubsection{ Scenarios }
A number of possible scenarios are formulated according to the types of camera devices. These scenarios involve:
\begin{itemize}
  \item \emph{C1:} VW camera-1 only.
  \item \emph{C2:} VW camera-2 only.
  \item \emph{C3:} NIR camera only.
  \item \emph{Ep1:} images of VW camera-1 constitute target set while images of VW camera-2 constitute query set. Both image sets are captured in the visible light.
  \item \emph{Ep2:} images of VW camera-1 constitute target set while images of NIR camera constitute query set.
  \item \emph{Ep3:} images of VW camera-2 constitute target set while images of VW camera-2 constitute query set.
\end{itemize}

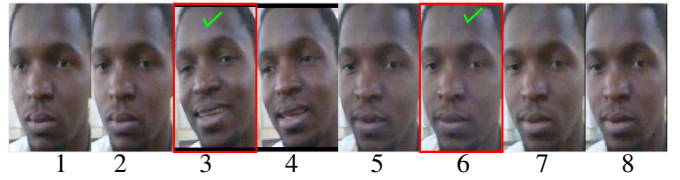
\begin{figure}[!]
  \centering
  \begin{overpic}[width=0.99\linewidth, height=2cm]{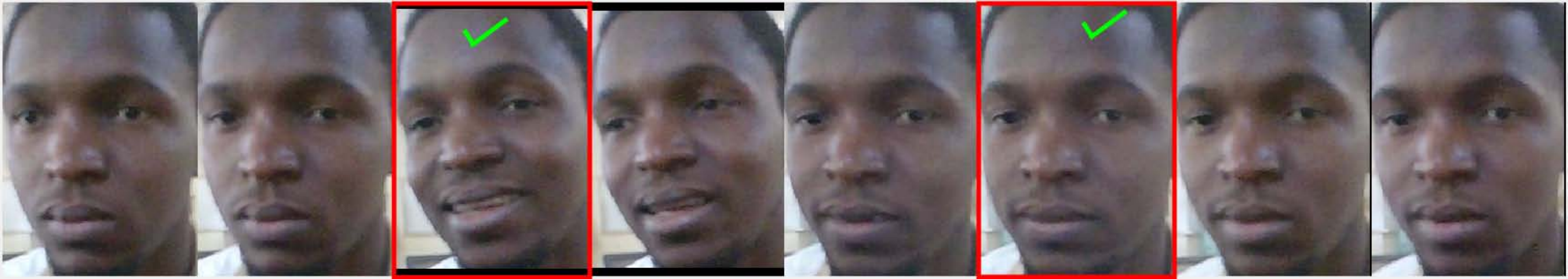}
  \put (7,-3) {1}
  \put (16,-3) {2}
  \put (29,-3) {3}
  \put (42,-3) {4}
  \put (55,-3) {5}
  \put (68,-3) {6}
  \put (80,-3) {7}
  \put (93,-3) {8}
  \end{overpic}
  \caption{ An example of image selection for the testing partition, No.3 and 6 images are maintained while others are threw away.}
 \label{fig:desimilariy_pic}\vspace{-3pt}
\end{figure}

In the aforementioned protocols, all images of one subject in the training partition will be utilized while \textcolor{\CorrectionTextColor}{only some images belonging to a subject are considered} in the testing partition. This is because, the face images of most subjects were captured in a single session within a short duration (few seconds) as a result of practical limitations. Hence, such images may resemble one another if facial movements of the subject are not obvious. To enlarge the intra-class variations and avoid duplicate results, face images from the same subject that are extremely similar to each other will be ignored and only one of them is considered in the testing evaluation protocol. Specifically, two images from the same subject are considered extremely similar if the computed correlation coefficient between them is \textcolor{\CorrectionTextColor}{not less than 0.8. By this criteria, only few distinctive images are selected} per subject in the testing partition as shown in \figref{fig:desimilariy_pic}. It can be observed that, only two images (No. 3 and 6) are maintained while the other images are ignored.
%The computation of similarity matrix in this section can be found in the \textcolor{\CorrectionTextColor}{supplementary materials. }
\textcolor{\CorrectionTextColor}{The decision to reduce some of the correlated images in testing was made to ensure algorithms are properly evaluated on the dataset. However, for training task, correlated images can still be useful. This is because, faces are composed of small components that are spatially positioned locally. In the event of little changes in one of these components, for instance, when a person smiles, even though holistically, the two faces (with and without smile) might be highly correlated, they will both be relevant during training, while for testing task (other than facial expression), this might produce redundant results.}

The details of the evaluation protocols for a variety of tasks involving training, classification, verification and identification are shown in \tabref{tab:protocols} and \tabref{tab:protocols_verification} respectively.

\begin{table}[]
\caption{Details of Evaluation Protocols.}
\renewcommand{\arraystretch}{1.1}   \setlength\tabcolsep{3pt}
\label{tab:protocols}
\centering
\begin{tabular}{|l|l|l|l|l|l|l|}
\hline
\multirow{2}{*}{Protocol} & \multicolumn{2}{c|}{C1} & \multicolumn{2}{c|}{C2} & \multicolumn{2}{c|}{C3} \\ \cline{2-7}
                          & Subjs      & Imgs       & Subjs      & Imgs       & Subjs      & Imgs       \\ \hline
\multicolumn{7}{|c|}{\textbf{Training}}                                                                          \\ \hline
ID-T-All-                 & 1110       & 15715      & 1131       & 15698      & 1133       & 14446      \\ \hline
ID-T-Split-               & 300        & 4122       & 300        & 4129       & 300        & 3767       \\ \hline
Expression-T-All-         & 77         & 1685       & 79         & 1657       & 79         & 1458       \\ \hline
\textcolor{\CorrectionTextColor}{Sex-T-All-}             & 1100       & 15610      & 1117       & 15461      & 1118       & 14225      \\ \hline
\textcolor{\CorrectionTextColor}{Sex-T-Split-}           & 440        & 6007       & 447        & 6268       & 447        & 5898       \\ \hline
Age-T-All-                & 985        & 14030      & 1002       & 13982      & 1003       & 12902      \\ \hline
Age-T-Split-              & 394        & 5616       & 401        & 5754       & 401        & 5159       \\ \hline
Ethnic-T-All-             & 500        & 7339       & 501        & 7396       & 503        & 6986       \\ \hline
Ethnic-T-Split-           & 200        & 2824       & 200        & 3019       & 200        & 2625       \\ \hline
\multicolumn{7}{|c|}{\textbf{Classification}}                                                                    \\ \hline
\multirow{2}{*}{Protocol} & \multicolumn{2}{c|}{C1} & \multicolumn{2}{c|}{C2} & \multicolumn{2}{c|}{C3} \\ \cline{2-7}
                          & Subjs      & Imgs       & Subjs      & Imgs       & Subjs      & Imgs       \\ \hline
Expression-C-All-         & 76         & 325        & 78         & 275        & 77         & 77         \\ \hline
\textcolor{\CorrectionTextColor}{Sex-C-All-}             & 1029       & 2437       & 1117       & 2440       & 1117       & 1180       \\ \hline
\textcolor{\CorrectionTextColor}{Sex-C-Split-}           & 613        & 1488       & 670        & 1477       & 670        & 703        \\ \hline
Age-C-All-                & 920        & 2237       & 1002       & 2202       & 1002       & 1061       \\ \hline
Age-C-Split-              & 548        & 1314       & 601        & 1309       & 602        & 634        \\ \hline
Ethnic-C-All-             & 466        & 994        & 501        & 1064       & 503        & 516        \\ \hline
Ethnic-C-Split-           & 275        & 579        & 301        & 615        & 303        & 317        \\ \hline
\multicolumn{7}{|c|}{\textbf{Close Identification}}                                                     \\ \hline
\multirow{2}{*}{Protocol} & \multicolumn{3}{c|}{Gallery}          & \multicolumn{3}{c|}{Probe}           \\ \cline{2-7}
                          & grp        & Subjs      & Imgs       & grp        & Subjs      & Imgs       \\ \hline
ID-I-All-Ep1              & C1         & 1038       & 1038       & C2         & 1036       & 2308       \\ \hline
ID-I-All-Ep2              & C1         & 1038       & 1038       & C3         & 1037       & 1097       \\ \hline
ID-I-All-Ep3              & C2         & 1131       & 1131       & C3         & 1130       & 1193       \\ \hline
ID-I-Split-Ep1            & C1         & 555        & 555        & C2         & 553        & 1213       \\ \hline
ID-I-Split-Ep2            & C1         & 555        & 555        & C3         & 554        & 587        \\ \hline
ID-I-Split-Ep3            & C2         & 602        & 602        & C3         & 601        & 635        \\ \hline

\multicolumn{7}{|c|}{\textbf{\textcolor{\CorrectionTextColorSecond}{Open Identification}}}                                                     \\ \hline
\multirow{2}{*}{\textcolor{\CorrectionTextColorSecond}{Protocol}} & \multicolumn{3}{c|}{\textcolor{\CorrectionTextColorSecond}{Gallery}}          & \multicolumn{3}{c|}{\textcolor{\CorrectionTextColorSecond}{Probe}}           \\ \cline{2-7}
                          & \textcolor{\CorrectionTextColorSecond}{grp}        & \textcolor{\CorrectionTextColorSecond}{Subjs}      & \textcolor{\CorrectionTextColorSecond}{Imgs}       & \textcolor{\CorrectionTextColorSecond}{grp}        & \textcolor{\CorrectionTextColorSecond}{Subjs}      & \textcolor{\CorrectionTextColorSecond}{Imgs}       \\
 \hline \textcolor{\CorrectionTextColorSecond}{ID-O-All-Ep1} & \textcolor{\CorrectionTextColorSecond}{C1} & \textcolor{\CorrectionTextColorSecond}{1038} & \textcolor{\CorrectionTextColorSecond}{2455} & \textcolor{\CorrectionTextColorSecond}{C2} & \textcolor{\CorrectionTextColorSecond}{1131} & \textcolor{\CorrectionTextColorSecond}{2466} \\
\textcolor{\CorrectionTextColorSecond}{ID-O-All-Ep2} & \textcolor{\CorrectionTextColorSecond}{C1} & \textcolor{\CorrectionTextColorSecond}{1038} & \textcolor{\CorrectionTextColorSecond}{2455} & \textcolor{\CorrectionTextColorSecond}{C3} & \textcolor{\CorrectionTextColorSecond}{1132} & \textcolor{\CorrectionTextColorSecond}{1195} \\
\textcolor{\CorrectionTextColorSecond}{ID-O-All-Ep3} & \textcolor{\CorrectionTextColorSecond}{C2} & \textcolor{\CorrectionTextColorSecond}{1131} & \textcolor{\CorrectionTextColorSecond}{2466} & \textcolor{\CorrectionTextColorSecond}{C3} & \textcolor{\CorrectionTextColorSecond}{1132} & \textcolor{\CorrectionTextColorSecond}{1195} \\ \hline
\textcolor{\CorrectionTextColorSecond}{ID-O-Split-Ep1} & \textcolor{\CorrectionTextColorSecond}{C1} & \textcolor{\CorrectionTextColorSecond}{540} & \textcolor{\CorrectionTextColorSecond}{1306} & \textcolor{\CorrectionTextColorSecond}{C2} & \textcolor{\CorrectionTextColorSecond}{596} & \textcolor{\CorrectionTextColorSecond}{1303} \\
\textcolor{\CorrectionTextColorSecond}{ID-O-Split-Ep2} & \textcolor{\CorrectionTextColorSecond}{C1} & \textcolor{\CorrectionTextColorSecond}{540} & \textcolor{\CorrectionTextColorSecond}{1306} & \textcolor{\CorrectionTextColorSecond}{C3} & \textcolor{\CorrectionTextColorSecond}{597} & \textcolor{\CorrectionTextColorSecond}{630} \\
\textcolor{\CorrectionTextColorSecond}{ID-O-Split-Ep3} & \textcolor{\CorrectionTextColorSecond}{C2} & \textcolor{\CorrectionTextColorSecond}{596} & \textcolor{\CorrectionTextColorSecond}{1303} & \textcolor{\CorrectionTextColorSecond}{C3} & \textcolor{\CorrectionTextColorSecond}{597} & \textcolor{\CorrectionTextColorSecond}{630} \\
\hline
\end{tabular}
\end{table}

\begin{table}[]
\centering
\renewcommand{\arraystretch}{1.1}
\setlength\tabcolsep{3pt}
\caption{Details of Evaluation Protocols: Verification.}
\label{tab:protocols_verification}
\begin{tabular}{|l|l|l|l|l|l|l|l|l|}
\hline
\multirow{2}{*}{Protocol} & \multicolumn{3}{c|}{Target} & \multicolumn{3}{c|}{Query} & \multicolumn{2}{c|}{Pairs} \\ \cline{2-9}
                          & grp    & Subjs    & Imgs    & grp    & Subjs    & Imgs   & \textcolor{\CorrectionTextColor}{Mated}   & \textcolor{\CorrectionTextColor}{Non Mated}   \\ \hline
ID-V-All-Ep1              & C1     & 1038     & 2455    & C2     & 1108     & 2426   &  \textcolor{\CorrectionTextColor}{6702} & \textcolor{\CorrectionTextColor}{755968}  \\ \hline
\begin{tabular}[c]{@{}l@{}}ID-V-All-Ep1-\\ Balanced\end{tabular}     & C1     & 1038     & 2455    & C2     & 1108     & 2426   & \textcolor{\CorrectionTextColor}{6702} & \textcolor{\CorrectionTextColor}{6702} \\ \hline
ID-V-All-Ep2              & C1     & 1038     & 2455    & C3     & 1109     & 1171   & \textcolor{\CorrectionTextColor}{2682} & \textcolor{\CorrectionTextColor}{472789}  \\ \hline
\begin{tabular}[c]{@{}l@{}}ID-V-All-Ep2-\\ Balanced\end{tabular}     & C1     & 1038     & 2455    & C3     & 1109     & 1171   & \textcolor{\CorrectionTextColor}{2682} & \textcolor{\CorrectionTextColor}{2682}               \\ \hline
ID-V-All-Ep3              & C2     & 1131     & 2466    & C3     & 1130     & 1193   & \textcolor{\CorrectionTextColor}{2662} & \textcolor{\CorrectionTextColor}{483643}                \\ \hline
\begin{tabular}[c]{@{}l@{}}ID-V-All-Ep3-\\ Balanced\end{tabular}     & C2     & 1131     & 2466    & C3     & 1130     & 1193   & \textcolor{\CorrectionTextColor}{2662} & \textcolor{\CorrectionTextColor}{2662}                \\ \hline
ID-V-Split-Ep1            & C1     & 555      & 1307    & C2     & 584      & 1259   & \textcolor{\CorrectionTextColor}{3484} & \textcolor{\CorrectionTextColor}{219133}          \\ \hline
ID-V-Split-Ep2            & C1     & 555      & 1307    & C3     & 585      & 618    & \textcolor{\CorrectionTextColor}{1431} & \textcolor{\CorrectionTextColor}{140036}           \\ \hline
ID-V-Split-Ep3            & C2     & 602      & 1293    & C3     & 601      & 635    & \textcolor{\CorrectionTextColor}{1389} & \textcolor{\CorrectionTextColor}{138490}          \\ \hline
\end{tabular}
\end{table}

%Kind see appendix \ref{sec:protocols_detail_summary}  for more information and details of each of the defined evaluation protocols.

\section{Experiments and Baseline Performances}
\label{sec:baseline_performance}

\subsection{Pre-processing}
\begin{figure}[!]
  \centering
  \begin{overpic}[width=0.8\linewidth, height=3cm ]{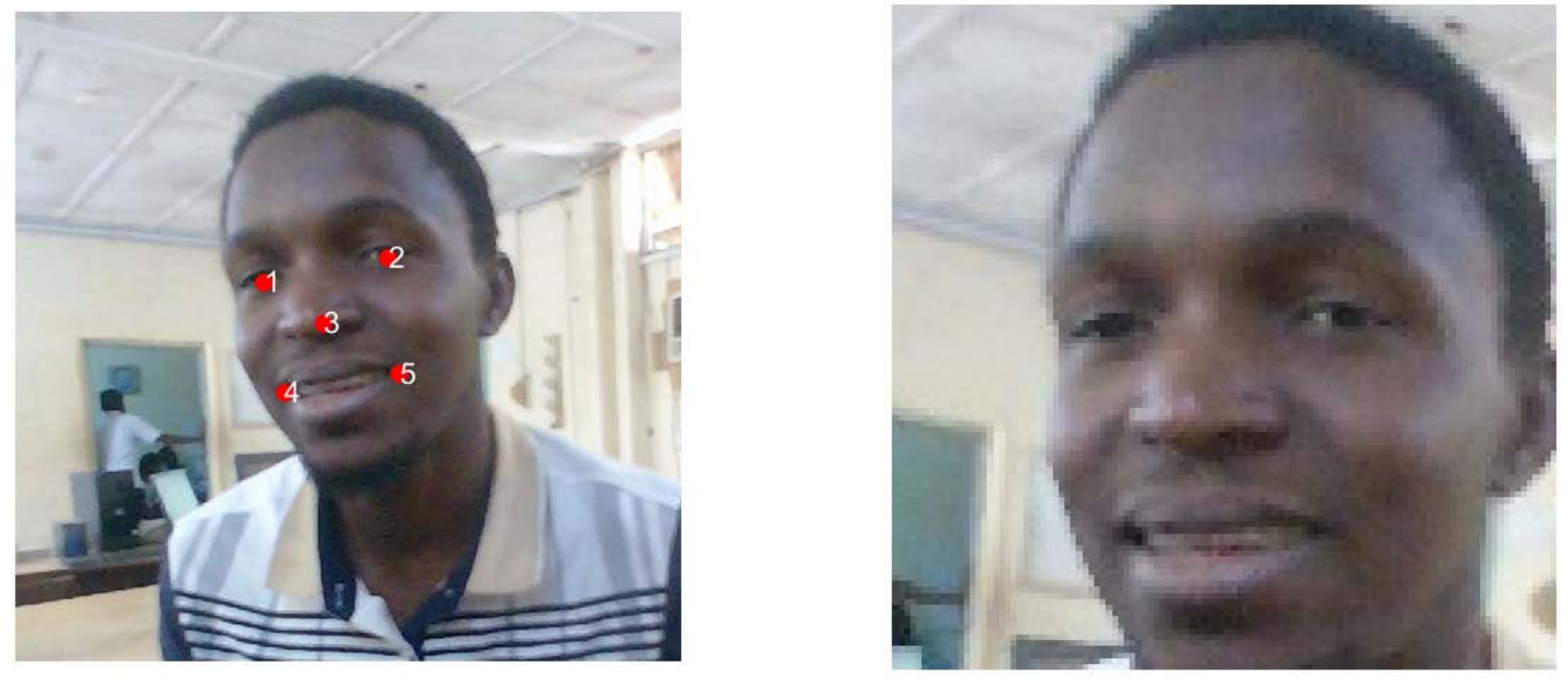}
  \end{overpic}
  \begin{overpic}[width=0.8\linewidth, height=3cm ]{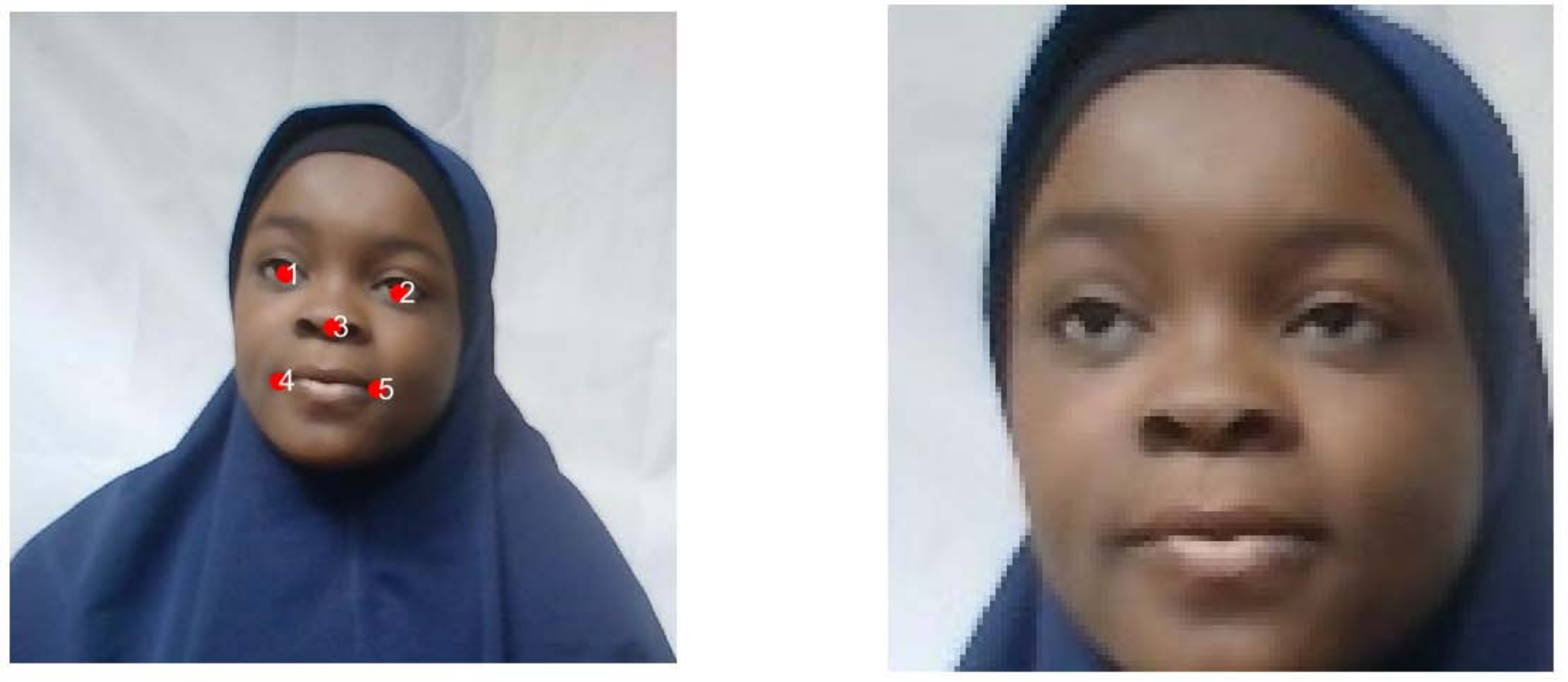}
  \put (16,-3) { \textcolor{\CorrectionTextColorSecond}{(a)}  }
  \put (72,-3) { \textcolor{\CorrectionTextColorSecond}{ (b) } }
  \end{overpic}
  \caption{Pre-processing of two face images in the database: \textcolor{\CorrectionTextColorSecond}{ (a) Original images with 5 landmarks; (b) Cropped and resized images } }
 \label{fig:alignment}\vspace{-5pt}
\end{figure}

Face images in the database \textcolor{\CorrectionTextColor}{were} aligned first and normalized before \textcolor{\CorrectionTextColor}{application of} face representation algorithms in some experiments. \textcolor{\CorrectionTextColorSecond}{ The pre-processing was adopted from \cite{Wang2018CosFace}. It involves the use of five facial landmarks (two eyes, nose and two mouth corners) to perform similarity transformation. Unlike in \cite{Wang2018CosFace}, in this work, the landmarks were manually labelled which makes the alignment more accurate. Based on the transformation co-efficients, the faces were then cropped and resized to $112\times 112$ . The pre-processing result is depicted in \figref{fig:alignment}. It can be observed that, despite the opposite head orientation of the subject, the relative position and scale of the face are roughly the same in the pre-processed image.}

\subsection{ Baseline Algorithms }
To evaluate the baseline performance on the database, the pre-trained models of some recent open-source SOTA face detection and recognition algorithms \textcolor{\CorrectionTextColor}{were} evaluated according to the prescribed protocols.
The \textcolor{\CorrectionTextColor}{baseline} face detection models are: S3FD\textsuperscript{e}\cite{zhang2017s3fd};  MTCNN\textsuperscript{f}\cite{zhang2016joint}; DSFD\textsuperscript{g}\cite{li2019dsfd}; and YOLOFace\textsuperscript{h} which are all trained with WIDER FACE ~\cite{yang2016wider} dataset. Note that MTCNN is additionally trained with CelebA~\cite{liu2015deep} dataset.

The feature extraction methods for face recognition include SphereFace\textsuperscript{a}\cite{Liu20} trained with CASIA-Webface, lightCNN variants\textsuperscript{b}\cite{Wu34} (lightCNN\-9, lightCNN-29 and lightCNN-29\_V2) trained with CASIA-Webface among which the lightCNN-29\_V2 model is additionally trained with  Ms-Celeb-1m\cite{Guo11},
ArcFace models\textsuperscript{c}\cite{Deng5}(ArcFace-model-r34-amf, ArcFace-model-r50-am-lfw and ArcFace-model-r100-ii) which are based on LResNet100E-IR network and trained with MS1MV2~\cite{Deng5}, CASIA-Arcface\textsuperscript{d} model trained with CASIA-Webface, CASIA-Softmax\textsuperscript{d} model trained with CASIA-Webface, Balanced-Softmax\textsuperscript{d} model trained with a balanced dataset containing \textcolor{\CorrectionTextColor}{an} equal number of all the races considered, Global-Softmax\textsuperscript{d} model trained with a dataset that resembles the ethnic distribution of the world population and MS1M-Arcface\textsuperscript{d} trained with MS1M-wo-RFW dataset.

\footnotetext{ The codes and models are downloaded from https://github.com at \textsuperscript{a}\href{https://github.com/wy1iu/sphereface}{/wy1iu/sphereface};  \textsuperscript{b}\href{https://github.com/AlfredXiangWu/LightCNN}{/AlfredXiangWu/LightCNN}; \textsuperscript{c}\href{https://github.com/deepinsight/insightface}{/deepinsight/insightface};     \textsuperscript{e}\href{https://github.com/weiliu89/caffe/tree/ssd}{/weiliu89/caffe/tree/ssd}; \textsuperscript{f}\href{https://github.com/kpzhang93/MTCNN_face_detection_alignment}{/kpzhang93/MTCNN\_face\_detection\_alignment};     \textsuperscript{g}\href{https://github.com/Tencent/FaceDetection-DSFD}{/Tencent/FaceDetection-DSFD};     \textsuperscript{h}\href{https://github.com/sthanhng/yoloface}{/sthanhng/yoloface}
and from \textsuperscript{d}\href{http://www.whdeng.cn/RFW/model.html}{http://www.whdeng.cn/RFW/model.html} }

\subsection{Experimental Results of Face Detection}
\begin{figure}[!tb]
  \centering
  \begin{overpic}[width=\linewidth]{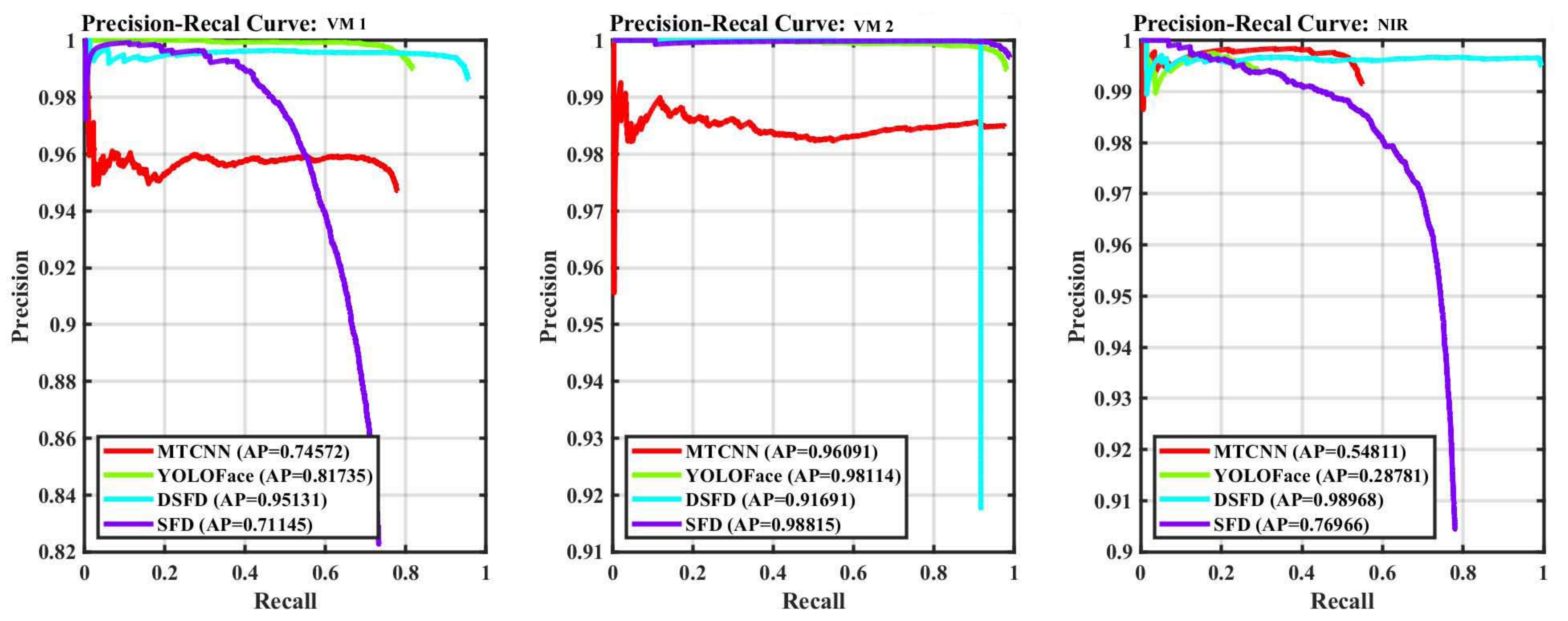}
  \put (14,-3){ (a) }
  \put (53,-3) { (b) }
  \put (80,-3) { (c) }
  \end{overpic}\vspace{4pt}
  \caption{Precision-recall curve and average precision of face detection algorithms on (a) C1 (b) C2 and (c) C3. }
 \label{fig:precision_recall}\vspace{-15pt}
\end{figure}

The experimental results of the face detection methods are presented in \figref{fig:precision_recall}. The precision-recall (PR) curves together with their corresponding average precisions are shown. It can be observed that the face detection results are almost saturated on the subset of C2 (VW camera 2), indicating little or no challenges. It is evident that C2 category mainly contains frontal faces with little pose and illumination variations. However, for the C1 (VW camera 1) and C3 (NIR camera) detection results, only DSFD \cite{li2019dsfd} is relatively successful in detecting the faces while the other algorithms perform poorly. This may be the consequences of  scale and illumination variations in C1 and the cross-spectrum discrepancy in C3.

\subsection{Experimental Results of Face Recognition}

\subsubsection{ Verification results at various \textcolor{\CorrectionTextColor}{FMRs} }
\begin{figure}
\vspace{2pt}
  \centering
  \begin{overpic}[width=\linewidth ]{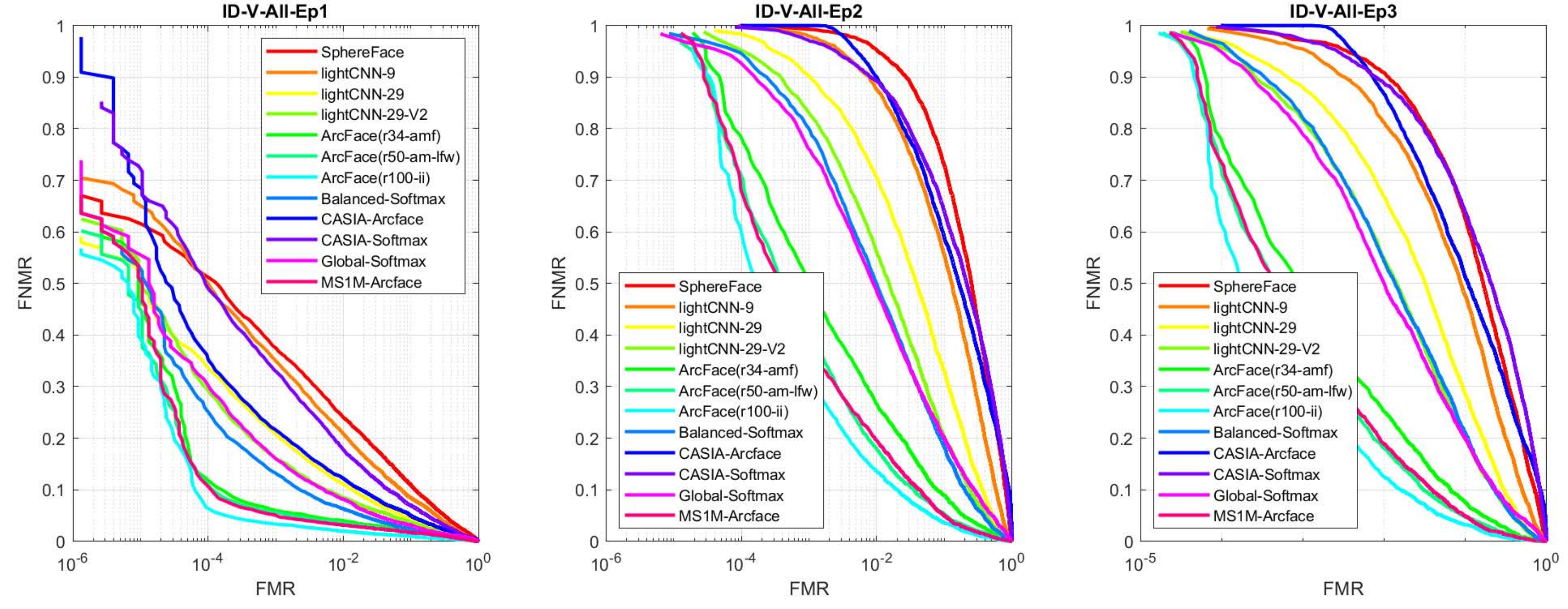}
  %\put (14,22) { (a) }
  % \put (53,22) { (b) }
  %\put (80,22) { (c) }
  \put (14,-3) { (a) }
  \put (53,-3) { (b) }
  \put (80,-3) { (c) }
  \end{overpic}
  \caption{ \textcolor{\CorrectionTextColorSecond}{DET curves of the verification results under the protocols: (a) ID-V-All-Ep1, (b) ID-V-All-Ep2, (c) ID-V-All-Ep3.} } %(d) ID-V-Split-Ep1, (e)ID-V-Split-Ep2, and (f) ID-V-Split-Ep3%  }
 \label{fig:verification_ROC_Curves}\vspace{-3pt}
\end{figure}

The verification results at various FMRs are presented in  \tabref{tab:verification_ID_V_All}. It can be observed that the overall performances across the baseline algorithms are relatively poor. \textcolor{\CorrectionTextColorSecond}{The ArcFace based models gains better results over other methods signifying higher feature discriminations against all the other models}. The \textcolor{\CorrectionTextColor}{DET curves} in \figref{fig:verification_ROC_Curves} also clarify the weak performance of the baseline algorithms and the superiority of ArcFace based models.

\begin{table*}[]
\centering
\renewcommand{\arraystretch}{1.1}
\setlength\tabcolsep{3pt}
\caption{ \textcolor{\CorrectionTextColorSecond}{Verification results: FNMR at various values of FMR under ID-V-All- Protocols  With and Without alignment.} }
\label{tab:verification_ID_V_All}

\begin{tabular}{|>{\color{\CorrectionTextColorSecond}}c|>{\color{\CorrectionTextColorSecond}}c|>{\color{\CorrectionTextColorSecond}}c|>{\color{\CorrectionTextColorSecond}}c|>{\color{\CorrectionTextColorSecond}}c|>{\color{\CorrectionTextColorSecond}}c|>{\color{\CorrectionTextColorSecond}}c|>{\color{\CorrectionTextColorSecond}}c|>{\color{\CorrectionTextColorSecond}}c|>{\color{\CorrectionTextColorSecond}}c|>{\color{\CorrectionTextColorSecond}}c|>{\color{\CorrectionTextColorSecond}}c|>{\color{\CorrectionTextColorSecond}}c|>{\color{\CorrectionTextColorSecond}}c|>{\color{\CorrectionTextColorSecond}}c|>{\color{\CorrectionTextColorSecond}}c|>{\color{\CorrectionTextColorSecond}}c|>{\color{\CorrectionTextColorSecond}}c|>{\color{\CorrectionTextColorSecond}}c|}
\hline
\multirow{3}{*}{Algorithm} & \multicolumn{6}{c|}{ID-V-All-Ep1 (FNMR@FMR)}    & \multicolumn{6}{c|}{ID-V-All-Ep2 (FNMR@FMR)}                                      & \multicolumn{6}{c|}{ID-V-All-Ep3 (FNMR@FMR)}
\\ \cline{2-19}
                           & \multicolumn{2}{c|}{0.01}     & \multicolumn{2}{c|}{0.001}       & \multicolumn{2}{c|}{0.0001}
                           & \multicolumn{2}{c|}{0.01}     & \multicolumn{2}{c|}{0.001}       & \multicolumn{2}{c|}{0.0001}
                           & \multicolumn{2}{c|}{0.01}     & \multicolumn{2}{c|}{0.001}       & \multicolumn{2}{c|}{0.0001}
                            \\ \cline{2-19}
                           & w/o & w             & w/o & w   & w/o & w           & w/o & w            & w/o & w             & w/o & w            & w/o & w      & w/o & w  & w/o & w     \\ \hline

ArcFace(r34-amf)       & 0.675 & 0.040 & 0.797 & 0.060 & 0.870 & 0.120 & 0.992 & 0.266 & 1.000 & 0.502 & 1.000 & 0.782 & 0.977 & 0.252 & 0.997 & 0.489 & 0.999 & 0.779 \\
ArcFace(r50-am-lfw)    & 0.706 & 0.035 & 0.813 & 0.056 & 0.883 & 0.112 & 0.991 & 0.179 & 0.999 & 0.397 & 1.000 & 0.713 & 0.973 & 0.181 & 0.995 & 0.394 & 1.000 & 0.709 \\
ArcFace(r100-ii)       & 0.657 & 0.019 & 0.769 & 0.033 & 0.845 & 0.065 & 0.968 & 0.136 & 0.994 & 0.310 & 0.998 & 0.598 & 0.953 & 0.125 & 0.987 & 0.317 & 0.997 & 0.616 \\ \hline
SphereFace             & 0.732 & 0.240 & 0.839 & 0.375 & 0.903 & 0.512 & 1.000 & 0.953 & 1.000 & 0.993 & 1.000 & 0.998 & 1.000 & 0.908 & 1.000 & 0.972 & 1.000 & 0.994 \\ \hline
lightCNN-29-V2         & 0.176 & 0.085 & 0.273 & 0.161 & 0.394 & 0.294 & 0.680 & 0.560 & 0.882 & 0.830 & 0.971 & 0.954 & 0.672 & 0.549 & 0.877 & 0.817 & 0.971 & 0.956 \\
lightCNN-29            & 0.259 & 0.111 & 0.370 & 0.209 & 0.494 & 0.336 & 0.840 & 0.707 & 0.961 & 0.903 & 0.995 & 0.984 & 0.806 & 0.668 & 0.946 & 0.881 & 0.990 & 0.969 \\
lightCNN-9             & 0.482 & 0.209 & 0.625 & 0.350 & 0.729 & 0.500 & 0.992 & 0.876 & 0.999 & 0.979 & 1.000 & 0.999 & 0.977 & 0.812 & 0.998 & 0.954 & 1.000 & 0.990 \\ \hline
Balanced-Softmax       & 0.686 & 0.064 & 0.832 & 0.132 & 0.916 & 0.250 & 0.991 & 0.500 & 1.000 & 0.801 & 1.000 & 0.940 & 0.982 & 0.544 & 0.998 & 0.821 & 1.000 & 0.966 \\
CASIA-Arcface          & 0.977 & 0.122 & 0.995 & 0.216 & 0.999 & 0.358 & 1.000 & 0.901 & 1.000 & 1.000 & 1.000 & 1.000 & 1.000 & 0.867 & 1.000 & 1.000 & 1.000 & 1.000 \\
CASIA-Softmax          & 0.810 & 0.177 & 0.917 & 0.330 & 0.964 & 0.491 & 1.000 & 0.892 & 1.000 & 0.972 & 1.000 & 0.997 & 0.998 & 0.889 & 1.000 & 0.973 & 1.000 & 0.997 \\
Global-Softmax         & 0.778 & 0.082 & 0.893 & 0.160 & 0.951 & 0.304 & 0.990 & 0.484 & 0.999 & 0.761 & 1.000 & 0.929 & 0.977 & 0.500 & 0.997 & 0.789 & 0.999 & 0.947 \\
MS1M-Arcface           & 0.769 & 0.031 & 0.874 & 0.051 & 0.931 & 0.107 & 0.997 & 0.196 & 1.000 & 0.383 & 1.000 & 0.680 & 0.992 & 0.189 & 1.000 & 0.399 & 1.000 & 0.733 \\ \hline

\end{tabular}
\end{table*}

\subsubsection{\textcolor{\CorrectionTextColorSecond}{Open Set Identification Results} }
\textcolor{\CorrectionTextColorSecond}{The results for the Detection and Identification Rates (DIRs) at 0.01 FMR for all subjects under the prescribed open set identification protocols are presented in \tabref{tab:open_identification_ID_I_All}. It can be observed that, for the visible wavelengths homogeneous protocol (Ep1), the performance is relative the same across the baseline models. However, for the two heterogeneous protocols (Ep2 and Ep3), the performance of the ArcFace loss based models is much higher.}

\begin{table}[]
\centering
\renewcommand{\arraystretch}{1.1}
\setlength\tabcolsep{3pt}
\caption{ \textcolor{\CorrectionTextColor}{DIRs@0.01FMR in (\%) under ID-I-All- protocols With and Without alignment.} }
\label{tab:open_identification_ID_I_All}
\begin{tabular}{|>{\color{\CorrectionTextColorSecond}}c|>{\color{\CorrectionTextColorSecond}}c|>{\color{\CorrectionTextColorSecond}}c|>{\color{\CorrectionTextColorSecond}}c|>{\color{\CorrectionTextColorSecond}}c|>{\color{\CorrectionTextColorSecond}}c|>{\color{\CorrectionTextColorSecond}}c|>{\color{\CorrectionTextColorSecond}}c|}
\hline
\multirow{3}{*}{Algorithm} & \multicolumn{2}{c|}{ID-I-All-Ep1}                         & \multicolumn{2}{c|}{ID-I-All-Ep2}                         & \multicolumn{2}{c|}{ID-I-All-Ep3}                         \\ \cline{2-7}
                           %& \multicolumn{2}{c|}{rank-1} & \multicolumn{2}{c|}{rank-1} & \multicolumn{2}{c|}{rank-1} \\ \cline{2-7}
                           & w/o          & w            & w/o          & w            & w/o          & w            \\ \hline

    ArcFace(r34-amf) & 21.5 & 80.6 & 0.5 & 39.5 & 13.3 & 79.0 \\
    ArcFace(r50-am-lfw) & 22.2 & 76.3 & 0.5 & 63.3 & 10.9 & 88.5 \\
    ArcFace(r100-ii) & 31.4 & 81.6 & 1.2 & 50.9 & 16.0 & 87.5 \\ \hline
    SphereFace & 19.6 & 74.5 & 0.3 & 7.8 & 1.6 & 49.1 \\ \hline
    lightCNN-29-V2 & 78.7 & 83.9 & 5.8 & 20.2 & 58.2 & 73.3 \\
    lightCNN-29 & 69.7 & 84.3 & 6.2 & 17.3 & 45.4 & 53.5 \\
    lightCNN-9 & 49.3 & 78.4 & 1.5 & 7.7 & 23.5 & 44.7 \\ \hline
    Balanced-Softmax & 19.8 & 79.0 & 0.0 & 26.1 & 6.5 & 56.7 \\
    CASIA-Arcface & 0.1 & 59.2 & 0.0 & 0.0 & 0.3 & 43.6 \\
    CASIA-Softmax & 9.7 & 64.9 & 0.0 & 1.3 & 1.3 & 35.0 \\
    Global-Softmax & 8.6 & 71.0 & 0.1 & 14.5 & 2.1 & 62.2 \\
    MS1M-Arcface & 16.1 & 75.8 & 0.1 & 62.5 & 7.1 & 85.2 \\ \hline

\end{tabular}
\end{table}

\subsubsection{\textcolor{\CorrectionTextColorSecond}{Close Set }Identification Results }
\begin{figure}
\vspace{2pt}
  \centering
  \begin{overpic}[width=\linewidth ]{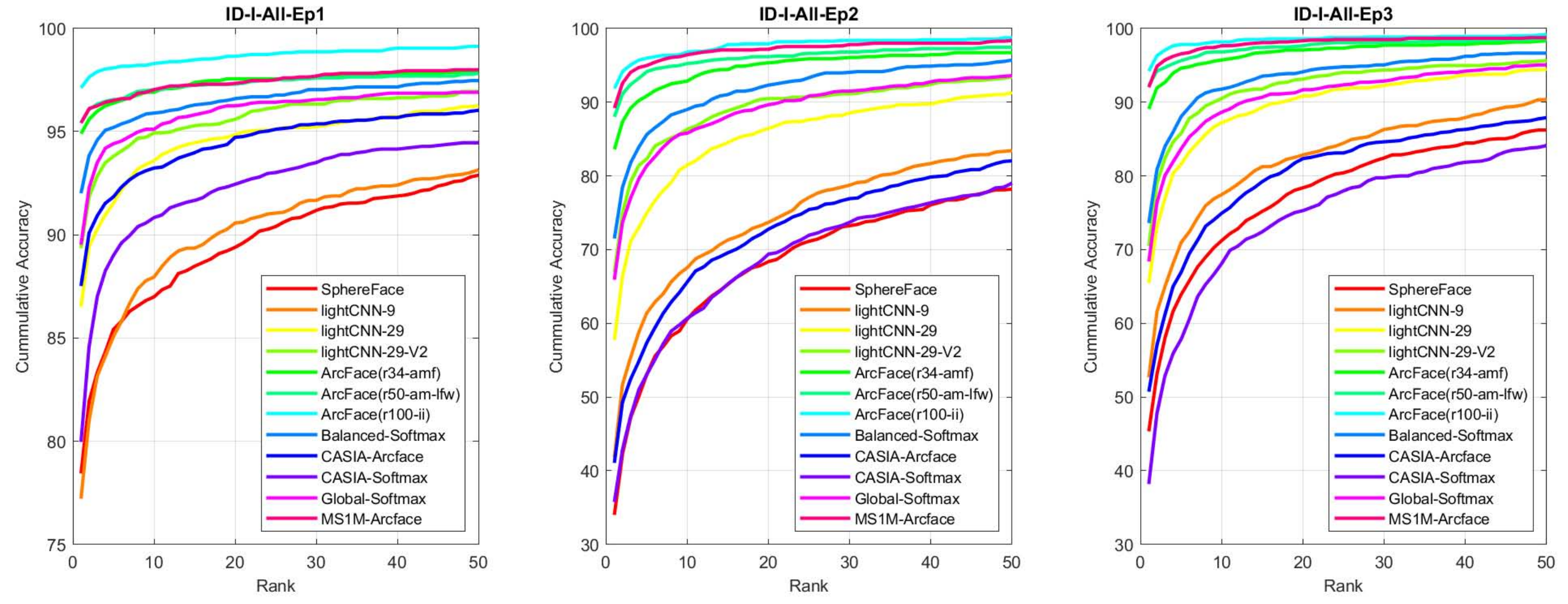}
  \put (14,-3) { (a) }
  \put (53,-3) { (b) }
  \put (80,-3) { (c) }
  \end{overpic}
  \caption{ \textcolor{\CorrectionTextColorSecond}{CMC curves of the 1:N identification results under the protocols: (a) ID-I-All-Ep1, (b) ID-I-All-Ep2, (c) ID-I-All-Ep3.} } %(d) ID-V-Split-Ep1, (e)ID-V-Split-Ep2, and (f) ID-V-Split-Ep3%  }
 \label{fig:CMC}\vspace{-3pt}
\end{figure}

The 1:N identification results for all subjects under the prescribed protocols are presented in \tabref{tab:identification_ID_I_All} \textcolor{\CorrectionTextColor}{ and the CMC shown in \figref{fig:CMC}}. \textcolor{\CorrectionTextColorSecond}{Like in the previous results, the ArcFace loss based} models outperform the other models in \textcolor{\CorrectionTextColor}{all the ranking} accuracy results. Also, the rank-1 accuracies are \textcolor{\CorrectionTextColorSecond}{relatively} low across the baseline algorithms. There are \textcolor{\CorrectionTextColorSecond}{substantial improvements with application of face alignment to the face images which signifies how sensitive the algorithms are to the pre-processing step.} % other results are reduced to some extent. The reason for \textcolor{\CorrectionTextColor}{these} discrepancies will be outlined in the subsequent section.
%More results can be found in the \textcolor{\CorrectionTextColor}{supplementary material attachments} like the results under ID-I-Split- protocols.

\begin{table*}[]
\centering
\renewcommand{\arraystretch}{1.1}
\setlength\tabcolsep{3pt}
\caption{ Close Identification results in (\%) under ID-I-All- protocols With and Without alignment.}
\label{tab:identification_ID_I_All}
\begin{tabular}{|>{\color{\CorrectionTextColorSecond}}c|>{\color{\CorrectionTextColorSecond}}c|>{\color{\CorrectionTextColorSecond}}c|>{\color{\CorrectionTextColorSecond}}c|>{\color{\CorrectionTextColorSecond}}c|>{\color{\CorrectionTextColorSecond}}c|
>{\color{\CorrectionTextColorSecond}}c|>{\color{\CorrectionTextColorSecond}}c|>{\color{\CorrectionTextColorSecond}}c|>{\color{\CorrectionTextColorSecond}}c|>{\color{\CorrectionTextColorSecond}}c|>{\color{\CorrectionTextColorSecond}}c|
>{\color{\CorrectionTextColorSecond}}c|>{\color{\CorrectionTextColorSecond}}c|>{\color{\CorrectionTextColorSecond}}c|>{\color{\CorrectionTextColorSecond}}c|>{\color{\CorrectionTextColorSecond}}c|>{\color{\CorrectionTextColorSecond}}c|>{\color{\CorrectionTextColorSecond}}c|}
\hline
\multirow{3}{*}{Algorithm} & \multicolumn{6}{c|}{ID-I-All-Ep1}                         & \multicolumn{6}{c|}{ID-I-All-Ep2}                         & \multicolumn{6}{c|}{ID-I-All-Ep3}                         \\ \cline{2-19}
                           & \multicolumn{2}{c|}{rank-1} & \multicolumn{2}{c|}{rank-5} & \multicolumn{2}{>{\color{\CorrectionTextColor}}c|}{rank-10} & \multicolumn{2}{c|}{rank-1} & \multicolumn{2}{c|}{rank-5} & \multicolumn{2}{>{\color{\CorrectionTextColor}}c|}{rank-10}  & \multicolumn{2}{c|}{rank-1} & \multicolumn{2}{c|}{rank-5} & \multicolumn{2}{c|}{rank-10} \\ \cline{2-19}
                           & w/o          & w            & w/o          & w            & w/o          & w            & w/o          & w            & w/o          & w            & w/o          & w           & w/o          & w            & w/o          & w   & w/o          & w \\ \hline

ArcFace(r34-amf) & 38.0 & 94.7 & 46.5 & 96.3 & 50.6 & 96.8 & 6.9 & 83.3 & 14.2 & 90.8 & 18.6 & 92.8 & 12.3 & 88.9 & 20.7 & 94.6 & 25.0 & 95.7 \\
ArcFace(r50-am-lfw) & 32.7 & 95.2 & 40.7 & 96.5 & 44.5 & 97.0 & 6.8 & 87.8 & 13.7 & 94.1 & 17.8 & 95.3 & 8.8 & 92.2 & 15.4 & 95.6 & 21.2 & 96.8 \\
ArcFace(r100-ii) & 36.8 & 96.9 & 44.9 & 98.0 & 48.1 & 98.2 & 11.8 & 91.7 & 18.7 & 96.1 & 23.1 & 96.9 & 13.1 & 94.1 & 23.2 & 97.8 & 28.3 & 98.2 \\ \hline
SphereFace & 32.5 & 83.7 & 40.6 & 89.6 & 44.8 & 91.3 & 2.3 & 42.5 & 6.1 & 60.6 & 8.9 & 68.2 & 3.4 & 53.3 & 6.5 & 71.8 & 9.5 & 78.3 \\ \hline
lightCNN-29-V2 & 80.4 & 91.0 & 85.6 & 95.0 & 87.5 & 95.7 & 54.6 & 71.6 & 70.4 & 84.4 & 75.5 & 88.3 & 56.7 & 75.6 & 73.4 & 88.6 & 79.5 & 92.3 \\
lightCNN-29 & 74.2 & 88.2 & 80.0 & 93.4 & 82.4 & 94.9 & 43.1 & 63.2 & 56.8 & 79.5 & 62.4 & 84.8 & 45.5 & 69.0 & 61.6 & 84.7 & 68.8 & 88.8 \\
lightCNN-9 & 57.6 & 80.7 & 67.5 & 87.4 & 71.5 & 90.2 & 17.1 & 47.2 & 29.8 & 65.0 & 36.1 & 71.8 & 20.3 & 57.8 & 34.9 & 75.9 & 41.8 & 80.9 \\ \hline
Balanced-Softmax & 34.1 & 91.8 & 45.7 & 95.0 & 50.4 & 95.7 & 6.2 & 71.3 & 13.7 & 85.6 & 19.3 & 88.9 & 6.9 & 73.4 & 15.3 & 88.0 & 19.7 & 91.8 \\
CASIA-Arcface & 27.5 & 87.3 & 35.7 & 91.7 & 39.5 & 93.1 & 0.3 & 41.1 & 1.0 & 57.4 & 1.8 & 65.4 & 0.4 & 50.6 & 1.4 & 67.0 & 1.9 & 74.9 \\
CASIA-Softmax & 23.1 & 79.8 & 33.7 & 88.9 & 38.9 & 90.7 & 1.5 & 35.8 & 3.6 & 53.1 & 5.9 & 60.6 & 3.2 & 38.1 & 6.6 & 57.8 & 9.6 & 68.2 \\
Global-Softmax & 20.9 & 89.3 & 31.0 & 94.3 & 36.1 & 95.0 & 1.5 & 65.8 & 5.6 & 81.4 & 8.8 & 85.8 & 3.3 & 68.3 & 8.5 & 83.7 & 12.2 & 88.7 \\
MS1M-Arcface & 28.8 & 95.2 & 37.1 & 96.4 & 41.4 & 97.0 & 3.9 & 89.0 & 9.1 & 95.1 & 12.5 & 96.5 & 7.0 & 91.9 & 13.9 & 96.6 & 18.6 & 97.7 \\ \hline

\end{tabular}
\end{table*}

\subsection{Analysis and Discussions}

\subsubsection{ Genuine and impostor Distribution  }
\begin{figure}[!tb]
  \centering
  \begin{overpic}[width=\linewidth ]{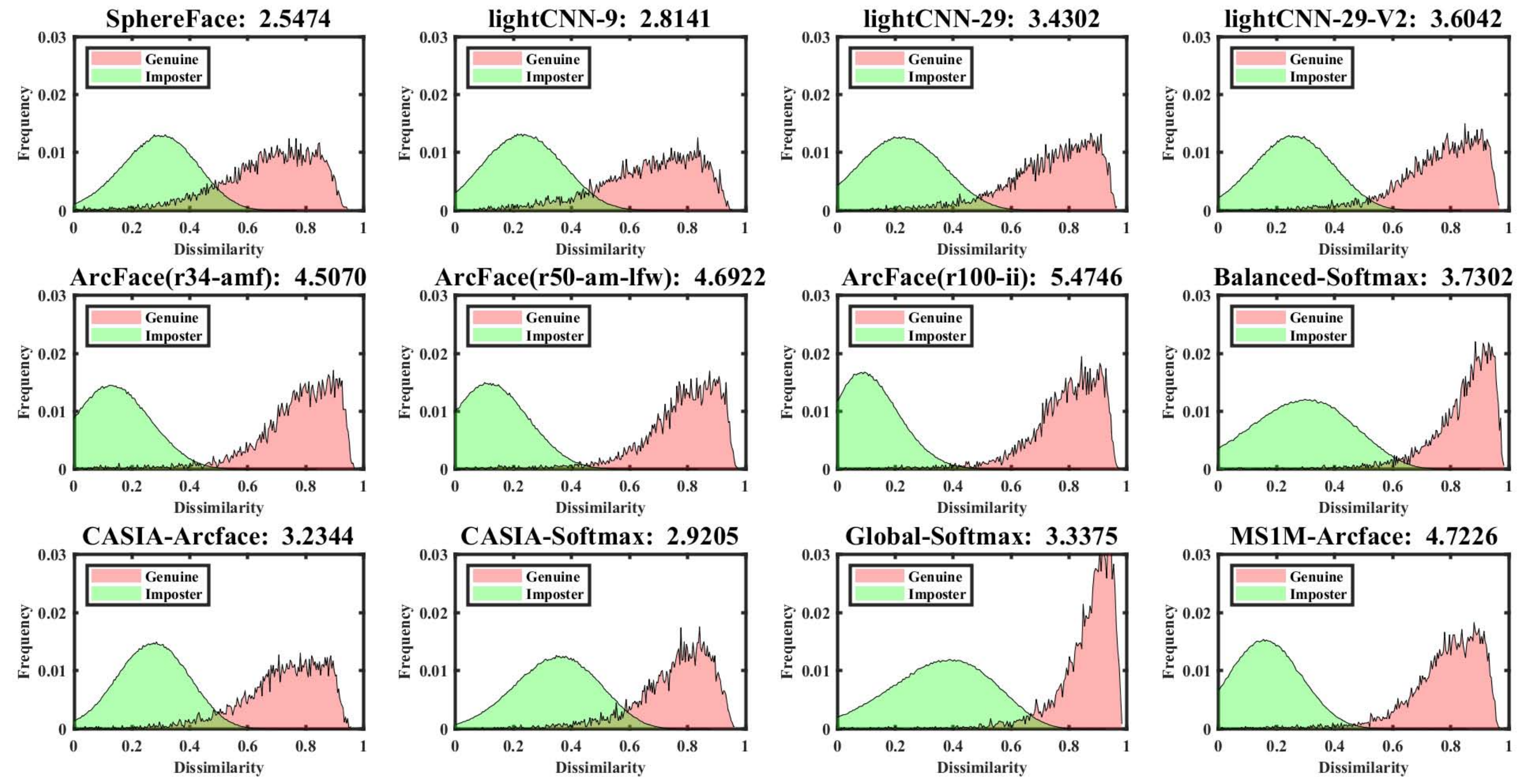}
  \end{overpic} %\vspace{1pt}
 \caption{ \textcolor{\CorrectionTextColorSecond}{DI values and genuine/impostor distributions of the baseline algorithms on aligned face images under ID-V-All-Ep1.}}
 \label{fig:features}
%  \vspace{-3pt}
\end{figure}

To analyze the properties of feature space, Decidability Index (DI) as suggested by Daugman\cite{daugman1998recognizing} is computed with the distribution of genuine and impostor pairs plotted as shown in \figref{fig:features} for the protocol ID-V-All-Ep1. This particular testing protocol was chosen here because of its simplicity as the images in the pairs are of the same spectrum (VW).  It can be observed that, the features extracted by the \textcolor{\CorrectionTextColorSecond}{ArcFace loss based} models (\figref{fig:features} (c),(d), and (e) ) are more separated than those extracted by the rest of the models with relatively high DI values. However, none of the models can pose enough separation for perfect classification of genuine or impostor images through a selected threshold. Undoubtedly, the proposed database exhibit many challenges.
%Note that more results of the other protocols can be found in the \textcolor{\CorrectionTextColor}{supplementary material attachments}.

\subsubsection{ Qualitative Analysis }
\begin{figure}
  \centering
  \begin{overpic}[width=\linewidth ]{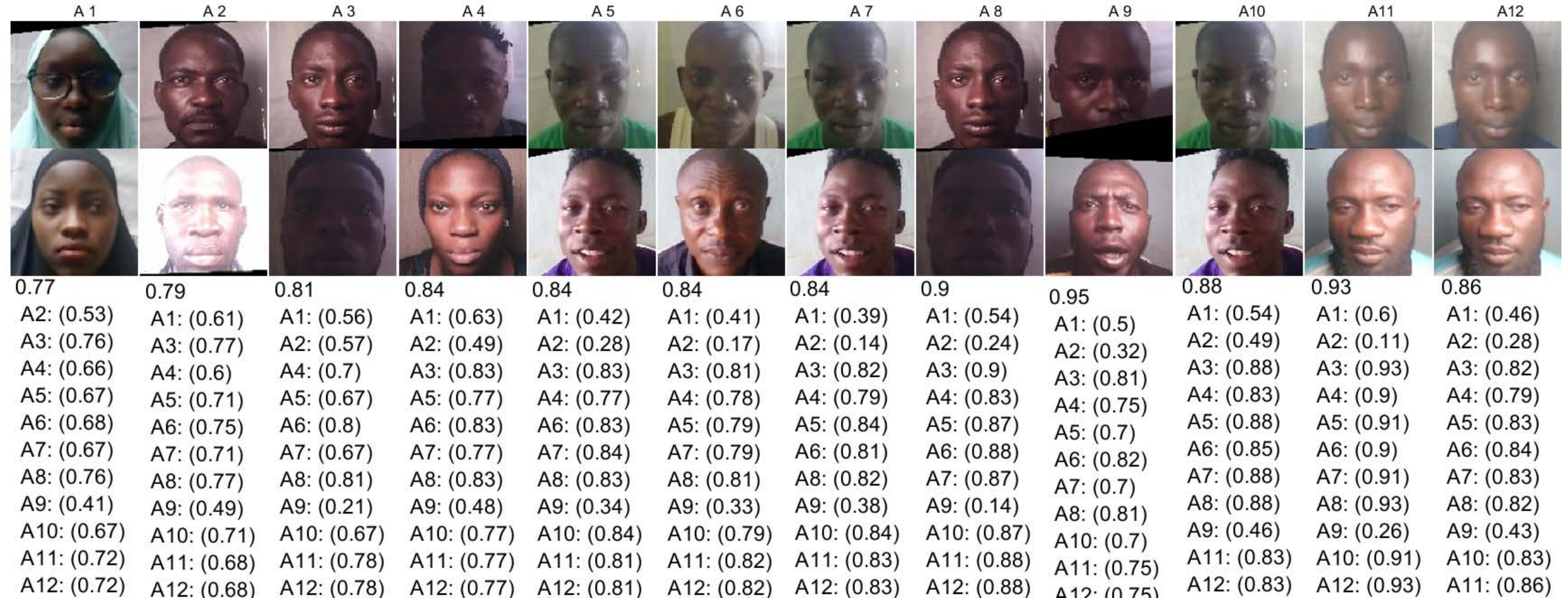}
  \end{overpic}
  \caption{ some selected challenging impostor pairs with similarity scores computed (scores without tag are for the corresponding algorithm): A1-\emph{SphereFace}, A2-\emph{LightCNN-9}, A3-\emph{LightCNN-29}, A4-\emph{LightCNN-29-V2}, A5-\emph{ArcFace-r34}, A6-\emph{ArcFace-r50},  A7-\emph{ArcFace-r100'},  A8-\emph{Balanced-Softmax}, A9-\emph{CASIA-Arcface}, A10-\emph{CASIA-Softmax}, A11-\emph{Global-Softmax}, A12-\emph{MS1M-Arcface}.  }
 \label{fig:worst_imposter}\vspace{-3pt}
\end{figure}

The \textcolor{\CorrectionTextColorSecond}{ArcFace loss based} models performs consistently better than the other methods across the experimental settings above. To further analyze the advantages of \textcolor{\CorrectionTextColorSecond}{these methods} on the proposed database, the matching scores of some impostor pairs using features extracted  \textcolor{\CorrectionTextColorSecond}{by the respective models} are visualized in \figref{fig:worst_imposter}. Apparently, the scores of \textcolor{\CorrectionTextColorSecond}{ArcFace loss based models are relatively lower than the other models. However, in general, it is incredible that despite the clarity of these images, the models still rate them with relatively higher similarity scores which can make them to be more susceptive to false acceptance. This further proves the inherent bias and ineffectiveness of these models on the proposed African subjects.}

\begin{figure}
\vspace{2pt}
  \centering
  \begin{overpic}[width=\linewidth ]{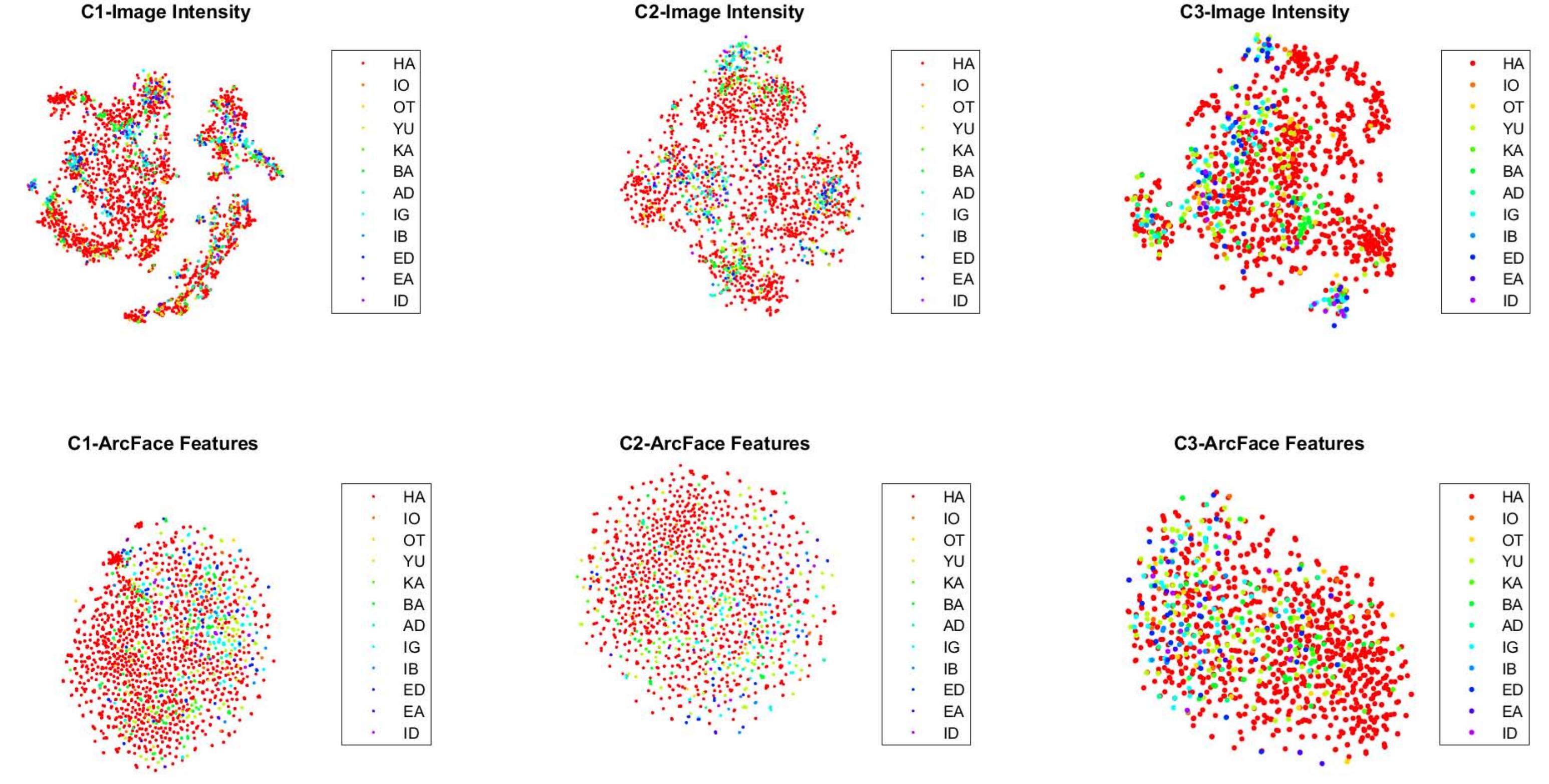}
  \put (14,-3) { (a) }
  \put (48,-3) { (b) }
  \put (80,-3) { (c) }
  \end{overpic}
  \caption{ \textcolor{\CorrectionTextColorSecond}{ The t-SNE visualisation of subject's face images according to their various ethnicities using image intensities (first row) and  extracted ArcFace(r100-ii) features (second row) of the camera categories (a) C1, (b) C2 and (c) C3.
  The meaning of abbreviations:  AD: Adamawa–Ubangi;   BA: Babur;   BT: Batangali;   EA: Eastern Nigeria;   ED: Edo;   HA: Hausa;   IB: Ibra;   ID: Idoma;   IG: Igala;   IO: Igbo;   KA: Kanuri;   NA: Nassarawa;   OT: Others;   YU: Yuruba  } }
 \label{fig:tsne_visualisation}\vspace{-3pt}
\end{figure}

\textcolor{\CorrectionTextColor}{Considering that there are many ethnicities in the database with some having a disproportionate number of subjects, it will be particularly significant to visualise the face images of these subjects according to their various ethinicities in a low dimensional plane. This will help in demonstrating the relative face variability of these ethnicities which can gauge the database representation among the ethnic groups. To achieve this task, we adopt the t-SNE data visualisation \cite{maaten2008visualizing} tool. We applied this tool on the images of the C1, C2 and C3 cameras using their intensity values and their respective facial features extracted by the  \textcolor{\CorrectionTextColorSecond}{ArcFace(r100-ii)}  model. The results are shown in  \figref{fig:tsne_visualisation}. It can be observed that, there are no clear boundaries of face variability across the ethnic groups which shows that the database can be an approximate representation of many other ethnic groups that are not currently captured in the database.}

\subsubsection{ The impact of face alignment }
\begin{figure}
\vspace{2pt}
  \centering
  \begin{overpic}[width=\linewidth ]{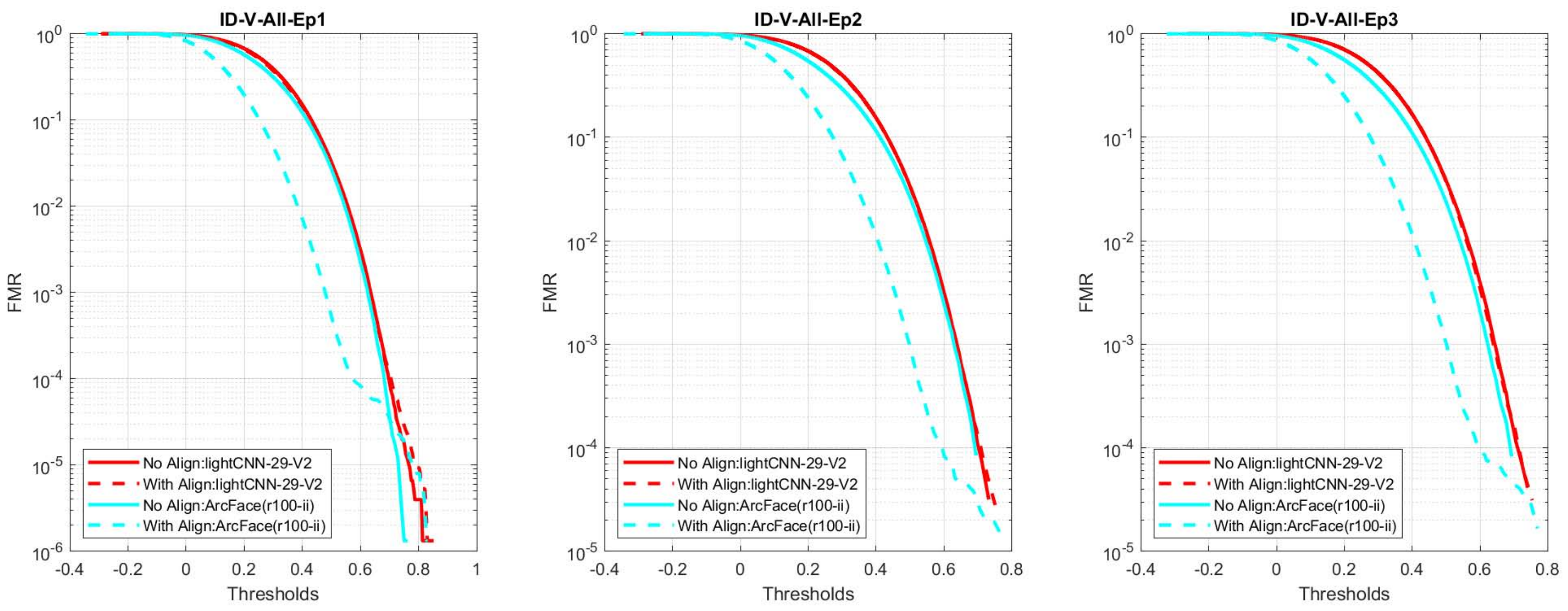}
  \put (14,-3) { (a) }
  \put (53,-3) { (b) }
  \put (80,-3) { (c) }
  \end{overpic}
  \caption{ \textcolor{\CorrectionTextColorSecond}{Impact of alignment on FMR (with and withOUT alignment) for the lightCNN-29-V2 and ArcFace(r100-ii) under the protocols: (a) ID-I-All-Ep1, (b) ID-I-All-Ep2, (c) ID-I-All-Ep3.} } %(d) ID-V-Split-Ep1, (e)ID-V-Split-Ep2, and (f) ID-V-Split-Ep3%  }
 \label{fig:FMR_curve}\vspace{-3pt}
\end{figure}

\begin{figure}[!tb]
  \centering
  \begin{overpic}[width=\linewidth ]{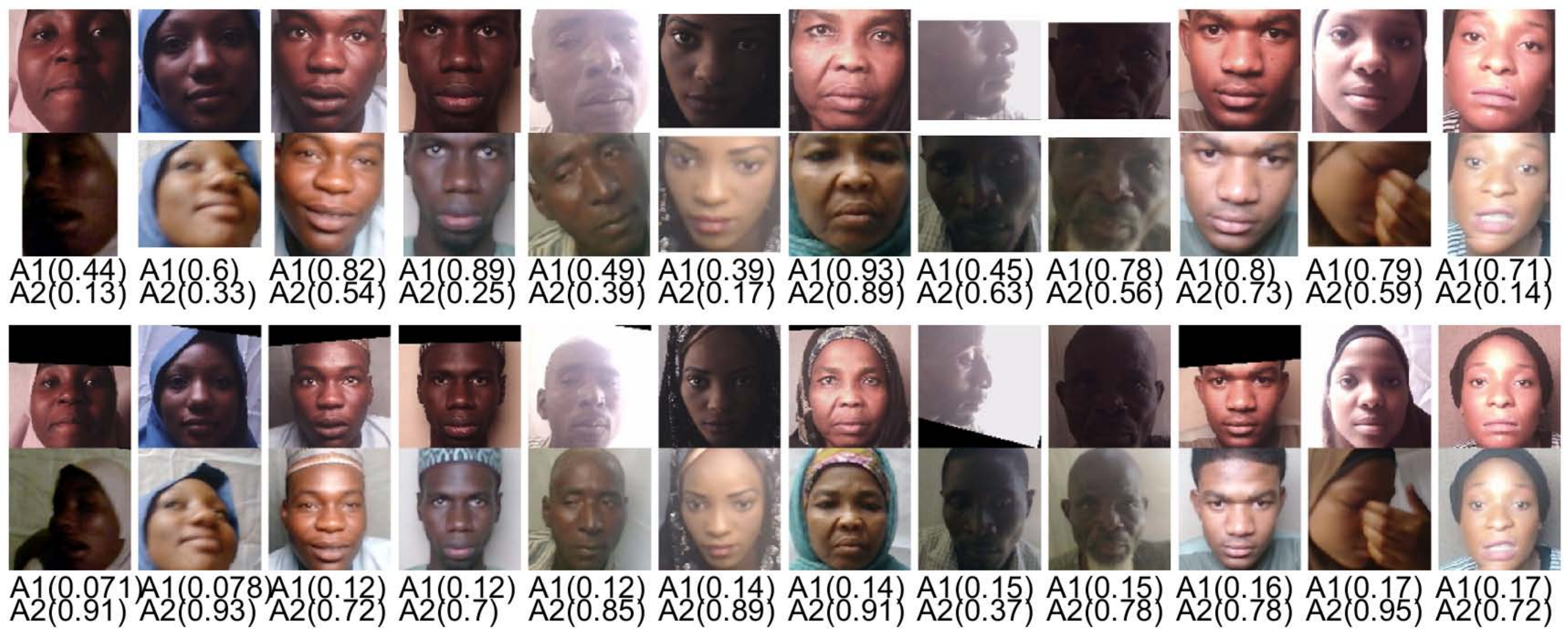}
  \put (-4,28) {(a)}
  \put (-4,10) {(b)}
  \end{overpic}
  \caption{ \textcolor{\CorrectionTextColorSecond}{Sample genuine pairs with their corresponding A1-lightCNN-29-V2 and A2-ArcFace(r100-ii) similarity scores: (a) without alignment and (b) with alignment.}  }
 \label{fig:genuine_alignment_problem}\vspace{-3pt}
\end{figure}

\begin{figure}[!tb]
  \centering
  \begin{overpic}[width=\linewidth ]{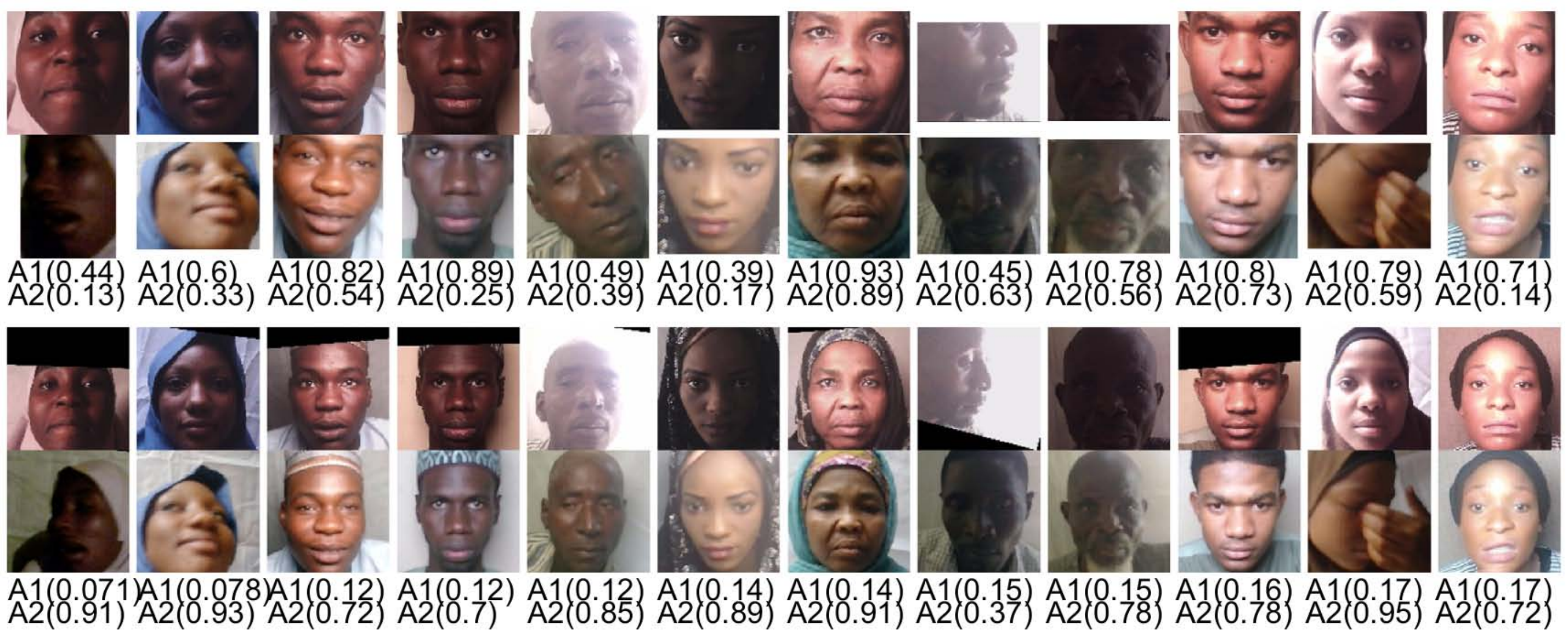}
  \put (-4,28) {(a)}
  \put (-4,10) {(b)}
  \end{overpic}
  \caption{ \textcolor{\CorrectionTextColorSecond}{Sample imposter pairs with their corresponding A1-lightCNN-29-V2 and A2-ArcFace(r100-ii) similarity scores: (a) without alignment and (b) with alignment.}  }
 \label{fig:imposter_alignment_problem}\vspace{-3pt}
\end{figure}

\textcolor{\CorrectionTextColorSecond}{Face alignment is considered to be an essential step before feature extraction which can generally improve face recognition performances. This is evident from the results presented in the previous section. However, for some models, the alignment on the proposed database (African subjects), is more effective than others. For ArcFace loss based models, the alignment substantially improve the results from near total failure to relatively low errors. However, for the lightCNN models the alignment is much less effective. This is evident from the  FMR plot of the lightCNN-29-V2 and ArcFace(r100-ii) models under various protocols as presented in \figref{fig:FMR_curve}. It can be observed that, at higher thresholds, the performance gap of FMR on the lightCNN-29-V2 model is almost the same with and without the alignment while the gap of the ArcFace(r100-ii) model is still wide. This can be visualised from the images of some genuine and imposter pairs with their corresponding lightCNN-29-V2 and ArcFace(r100-ii) similarity scores as shown in \figref{fig:genuine_alignment_problem} for genuine pairs and \figref{fig:imposter_alignment_problem} for the imposter pairs. It can be observed that, the images of which the African face physiognomic and topological details are reviled before the alignment are poorly graded by the ArcFace(r100-ii) despite having relatively better FMR results. This could be an evidence that the African subjects' FMR can be severely affected when the alignment is poorly performed due to inaccuracies of the detected landmark points used for the alignment.}

\section{Conclusions and Future Work}
\label{sec::conclusion}
This paper present a large-scale African face image database developed for \textcolor{\CorrectionTextColor}{mitigating} the racial bias problem in research and applications of face recognition. The database  \textcolor{\CorrectionTextColor}{was} collected from 1,183 African subjects using both visible and near-infrared cameras. And the face images are labeled with 68 landmark points to facilitate research of face image preprocessing and feature analysis. The evaluation protocols and baseline performances are also published with the database. The preliminary experiments show that SOTA face recognition methods in \textcolor{\CorrectionTextColor}{the} public domain can not achieve satisfying results on the database, which demonstrate the need for more studies on face recognition of African subjects.
%\textcolor{\CorrectionTextColor}{Although, the presented database was captured using multiple cameras, the intra- class variability in terms of imaging conditions is relatively }
Our future work will focus on developing better machine learning and pattern recognition methods for African face biometrics.

\bibliographystyle{IEEEtran}
\bibliography{IEEEabrv,CASIABlackFace}

\begin{thebibliography}{10}
\providecommand{\url}[1]{#1}
\csname url@samestyle\endcsname
\providecommand{\newblock}{\relax}
\providecommand{\bibinfo}[2]{#2}
\providecommand{\BIBentrySTDinterwordspacing}{\spaceskip=0pt\relax}
\providecommand{\BIBentryALTinterwordstretchfactor}{4}
\providecommand{\BIBentryALTinterwordspacing}{\spaceskip=\fontdimen2\font plus
\BIBentryALTinterwordstretchfactor\fontdimen3\font minus
  \fontdimen4\font\relax}
\providecommand{\BIBforeignlanguage}[2]{{%
\expandafter\ifx\csname l@#1\endcsname\relax
\typeout{** WARNING: IEEEtran.bst: No hyphenation pattern has been}%
\typeout{** loaded for the language `#1'. Using the pattern for}%
\typeout{** the default language instead.}%
\else
\language=\csname l@#1\endcsname
\fi
#2}}
\providecommand{\BIBdecl}{\relax}
\BIBdecl

\bibitem{Li18}
Z.~Li, Y.~Hu, R.~He, and Z.~Sun, ``Learning disentangling and fusing networks
  for face completion under structured occlusions,'' \emph{Pattern
  Recognition}, vol.~99, p. 107073, 2020.

\bibitem{Zhang36}
H.~Zhang, Q.~Li, and Z.~Sun, ``Adversarial learning semantic volume for 2d/3d
  face shape regression in the wild,'' \emph{IEEE Transactions on Image
  Processing}, vol.~28, no.~9, pp. 4526--4540, 2019.

\bibitem{Cao2}
J.~Cao, Y.~Hu, H.~Zhang, R.~He, and Z.~Sun, ``Towards high fidelity face
  frontalization in the wild,'' \emph{International Journal of Computer
  Vision}, pp. 1--20, 2019.

\bibitem{Li17}
Y.~Li, L.~Song, X.~Wu, R.~He, and T.~Tan, ``Anti-makeup: Learning a bi-level
  adversarial network for makeup-invariant face verification,'' in
  \emph{Thirty-Second AAAI Conference on Artificial Intelligence}, Conference
  Proceedings.

\bibitem{Liang19}
J.~Liang, R.~He, Z.~Sun, and T.~Tan, ``Distant supervised centroid shift: A
  simple and efficient approach to visual domain adaptation,'' in
  \emph{Proceedings of the IEEE Conference on Computer Vision and Pattern
  Recognition}, Conference Proceedings, pp. 2975--2984.

\bibitem{Li16}
P.~Li, X.~Wu, Y.~Hu, R.~He, and Z.~Sun, ``M2fpa: A multi-yaw multi-pitch
  high-quality dataset and benchmark for facial pose analysis,'' in
  \emph{Proceedings of the IEEE International Conference on Computer Vision},
  Conference Proceedings, pp. 10\,043--10\,051.

\bibitem{Robinson28}
J.~P. Robinson, M.~Shao, Y.~Wu, and Y.~Fu, ``Families in the wild (fiw)
  large-scale kinship image database and benchmarks,'' in \emph{Proceedings of
  the 24th ACM international conference on Multimedia}, Conference Proceedings,
  pp. 242--246.

\bibitem{Rossler29}
A.~Rossler, D.~Cozzolino, L.~Verdoliva, C.~Riess, J.~Thies, and M.~Niebner,
  ``Faceforensics: A large-scale video dataset for forgery detection in human
  faces,'' \emph{arXiv preprint arXiv:1803.09179}, 2018.

\bibitem{Liu21}
Y.~Liu, Q.~Li, and Z.~Sun, ``Attribute-aware face aging with wavelet-based
  generative adversarial networks,'' in \emph{Proceedings of the IEEE
  Conference on Computer Vision and Pattern Recognition}, Conference
  Proceedings, pp. 11\,877--11\,886.

\bibitem{Li15}
P.~Li, Y.~Hu, X.~Wu, R.~He, and Z.~Sun, ``Deep label refinement for age
  estimation,'' \emph{Pattern Recognition}, vol. 100, p. 107178, 2020.

\bibitem{Huang12}
G.~B. Huang, M.~Mattar, T.~Berg, and E.~Learned-Miller, ``Labeled faces in the
  wild: A database forstudying face recognition in unconstrained
  environments,'' Conference Proceedings.

\bibitem{Yi35}
D.~Yi, Z.~Lei, S.~Liao, and S.~Z. Li, ``Learning face representation from
  scratch,'' \emph{arXiv preprint arXiv:1411.7923}, 2014.

\bibitem{Cao3}
Q.~Cao, L.~Shen, W.~Xie, O.~M. Parkhi, and A.~Zisserman, ``Vggface2: A dataset
  for recognising faces across pose and age,'' in \emph{2018 13th IEEE
  International Conference on Automatic Face and Gesture Recognition (FG
  2018)}.\hskip 1em plus 0.5em minus 0.4em\relax IEEE, Conference Proceedings,
  pp. 67--74.

\bibitem{Guo11}
Y.~Guo, L.~Zhang, Y.~Hu, X.~He, and J.~Gao, ``Ms-celeb-1m: A dataset and
  benchmark for large-scale face recognition,'' in \emph{European conference on
  computer vision}.\hskip 1em plus 0.5em minus 0.4em\relax Springer, Conference
  Proceedings, pp. 87--102.

\bibitem{Kemelmacher13}
I.~Kemelmacher-Shlizerman, S.~M. Seitz, D.~Miller, and E.~Brossard, ``The
  megaface benchmark: 1 million faces for recognition at scale,'' in
  \emph{Proceedings of the IEEE conference on computer vision and pattern
  recognition}, Conference Proceedings, pp. 4873--4882.

\bibitem{Phillips24}
P.~J. Phillips, P.~J. Flynn, T.~Scruggs, K.~W. Bowyer, J.~Chang, K.~Hoffman,
  J.~Marques, J.~Min, and W.~Worek, ``Overview of the face recognition grand
  challenge,'' in \emph{2005 IEEE computer society conference on computer
  vision and pattern recognition (CVPR'05)}, vol.~1.\hskip 1em plus 0.5em minus
  0.4em\relax IEEE, Conference Proceedings, pp. 947--954.

\bibitem{Gao9}
W.~Gao, B.~Cao, S.~Shan, X.~Chen, D.~Zhou, X.~Zhang, and D.~Zhao, ``The
  cas-peal large-scale chinese face database and baseline evaluations,''
  \emph{IEEE Transactions on Systems, Man, and Cybernetics-Part A: Systems and
  Humans}, vol.~38, no.~1, pp. 149--161, 2007.

\bibitem{Milborrow22}
S.~Milborrow, J.~Morkel, and F.~Nicolls, ``The muct landmarked face database,''
  \emph{Pattern Recognition Association of South Africa}, vol. 201, no.~0,
  2010.

\bibitem{Patrick2019Face}
P.~Grother, N.~Mei, and H.~Kayee, ``Face recognition vendor test (frvt) part 3:
  Demographic effects,'' National Institute of Standards and Technology, Report
  NISTIR 8280, 2019.

\bibitem{Buolamwini1}
J.~Buolamwini and T.~Gebru, ``Gender shades: Intersectional accuracy
  disparities in commercial gender classification,'' in \emph{Conference on
  fairness, accountability and transparency}, Conference Proceedings, pp.
  77--91.

\bibitem{Du6}
M.~Du, F.~Yang, N.~Zou, and X.~Hu, ``Fairness in deep learning: A computational
  perspective,'' \emph{arXiv preprint arXiv:1908.08843}, 2019.

\bibitem{Wang33}
M.~Wang, W.~Deng, J.~Hu, X.~Tao, and Y.~Huang, ``Racial faces in the wild:
  Reducing racial bias by information maximization adaptation network,'' in
  \emph{Proceedings of the IEEE International Conference on Computer Vision},
  Conference Proceedings, pp. 692--702.

\bibitem{Wang32}
M.~Wang and W.~Deng, ``Mitigate bias in face recognition using skewness-aware
  reinforcement learning,'' \emph{arXiv preprint arXiv:1911.10692}, 2019.

\bibitem{Singer31}
\BIBentryALTinterwordspacing
N.~Singer, ``Amazon's facial recognition wrongly identifies 28 lawmakers,
  a.c.l.u. says,'' 2020-02-23 2018. [Online]. Available:
  \url{https://www.nytimes.com/2018/07/26/technology/amazon-aclu-facial-recognition-congress.html}
\BIBentrySTDinterwordspacing

\bibitem{Nagpal23}
S.~Nagpal, M.~Singh, R.~Singh, M.~Vatsa, and N.~Ratha, ``Deep learning for face
  recognition: Pride or prejudiced?'' \emph{arXiv preprint arXiv:1904.01219},
  2019.

\bibitem{cook2019demographic}
\BIBentryALTinterwordspacing
C.~M. Cook, J.~J. Howard, Y.~B. Sirotin, J.~L. Tipton, and A.~R. Vemury,
  ``Demographic {Effects} in {Facial} {Recognition} and {Their} {Dependence} on
  {Image} {Acquisition}: {An} {Evaluation} of {Eleven} {Commercial}
  {Systems},'' \emph{IEEE Transactions on Biometrics, Behavior, and Identity
  Science}, vol.~1, no.~1, pp. 32--41, Jan. 2019. [Online]. Available:
  \url{https://ieeexplore.ieee.org/document/8636231/}
\BIBentrySTDinterwordspacing

\bibitem{Howard2019The}
J.~J. Howard, Y.~B. Sirotin, and A.~R. Vemury, ``The effect of broad and
  specific demographic homogeneity on the imposter distributions and false
  match rates in face recognition algorithm performance,'' in \emph{2019 IEEE
  10th International Conference on Biometrics Theory, Applications and Systems
  (BTAS)}, 2019.

\bibitem{Krishnapriya2020Issues}
K.~S. Krishnapriya, V.~Albiero, K.~Vangara, M.~C. King, and K.~W. Bowyer,
  ``Issues related to face recognition accuracy varying based on race and skin
  tone,'' \emph{IEEE Transactions on Technology and Society}, vol.~1, no.~1,
  pp. 8--20, 2020.

\bibitem{Drozdowski2020Demographic}
P.~Drozdowski, C.~Rathgeb, A.~Dantcheva, N.~Damer, and C.~Busch, ``Demographic
  bias in biometrics: A survey on an emerging challenge,'' \emph{IEEE
  Transactions on Technology \& Society}, 2020.

\bibitem{Furl8}
N.~Furl, P.~J. Phillips, and A.~J. O'Toole, ``Face recognition algorithms and
  the other-race effect: computational mechanisms for a developmental contact
  hypothesis,'' \emph{Cognitive Science}, vol.~26, no.~6, pp. 797--815, 2002.

\bibitem{ISO201522116}
ISO, ``Information technology - a study of the differential impact of
  demographic factors in biometric recognition system performance,'' p.
  22116.2, 2020.

\bibitem{Cavazos4}
J.~G. Cavazos, P.~J. Phillips, C.~D. Castillo, and A.~J. O'Toole, ``Accuracy
  comparison across face recognition algorithms: Where are we on measuring race
  bias?'' \emph{arXiv preprint arXiv:1912.07398}, 2019.

\bibitem{Phillips25}
P.~J. Phillips, F.~Jiang, A.~Narvekar, J.~Ayyad, and A.~J. O'Toole, ``An
  other-race effect for face recognition algorithms,'' \emph{ACM Transactions
  on Applied Perception (TAP)}, vol.~8, no.~2, pp. 1--11, 2011.

\bibitem{un2015population}
UN, ``Population 2030: Demographic challenges and opportunities for sustainable
  development planning,'' 2015.

\bibitem{Azzopardi2016Gender}
G.~Azzopardi, A.~Greco, and M.~Vento, ``Gender recognition from face images
  with trainable cosfire filters,'' in \emph{IEEE International Conference on
  Advanced Video \& Signal Based Surveillance}, 2016.

\bibitem{Phillips26}
P.~J. Phillips, H.~Moon, S.~A. Rizvi, and P.~J. Rauss, ``The feret evaluation
  methodology for face-recognition algorithms,'' \emph{IEEE Transactions on
  pattern analysis and machine intelligence}, vol.~22, no.~10, pp. 1090--1104,
  2000.

\bibitem{morales2020sensitivenets}
A.~Morales, J.~Fierrez, R.~Vera-Rodriguez, and R.~Tolosana, ``Sensitivenets:
  Learning agnostic representations with application to face images,''
  \emph{IEEE Transactions on Pattern Analysis and Machine Intelligence}, 2020.

\bibitem{2019Diversity}
M.~Merler, N.~Ratha, R.~S. Feris, and J.~R. Smith, ``Diversity in faces,''
  \emph{arXiv preprint arXiv:1804.02767}, 2019.

\bibitem{2019FairFace}
K.~Krkkinen and J.~Joo, ``Fairface: Face attribute dataset for balanced race,
  gender, and age,'' 2019.

\bibitem{2017Learning}
Y.~Li, J.~Yang, Y.~Song, L.~Cao, and L.~J. Li, ``Learning from noisy labels
  with distillation,'' in \emph{2017 IEEE International Conference on Computer
  Vision (ICCV)}, 2017.

\bibitem{2019DemogPairs}
I.~Hupont and C.~Fernandez, ``Demogpairs: Quantifying the impact of demographic
  imbalance in deep face recognition,'' in \emph{2019 14th IEEE International
  Conference on Automatic Face \& Gesture Recognition (FG 2019)}, 2019.

\bibitem{Ricanek27}
K.~Ricanek and T.~Tesafaye, ``Morph: A longitudinal image database of normal
  adult age-progression,'' in \emph{7th International Conference on Automatic
  Face and Gesture Recognition (FGR06)}.\hskip 1em plus 0.5em minus 0.4em\relax
  IEEE, Conference Proceedings, pp. 341--345.

\bibitem{Klare14}
B.~F. Klare, B.~Klein, E.~Taborsky, A.~Blanton, J.~Cheney, K.~Allen,
  P.~Grother, A.~Mah, and A.~K. Jain, ``Pushing the frontiers of unconstrained
  face detection and recognition: Iarpa janus benchmark a,'' in
  \emph{Proceedings of the IEEE conference on computer vision and pattern
  recognition}, Conference Proceedings, pp. 1931--1939.

\bibitem{Eidinger7}
E.~Eidinger, R.~Enbar, and T.~Hassner, ``Age and gender estimation of
  unfiltered faces,'' \emph{IEEE Transactions on Information Forensics and
  Security}, vol.~9, no.~12, pp. 2170--2179, 2014.

\bibitem{yang2016wider}
S.~Yang, P.~Luo, C.-C. Loy, and X.~Tang, ``Wider face: A face detection
  benchmark,'' in \emph{Proceedings of the IEEE conference on computer vision
  and pattern recognition}, 2016, pp. 5525--5533.

\bibitem{moschoglou2017agedb}
S.~Moschoglou, A.~Papaioannou, C.~Sagonas, J.~Deng, I.~Kotsia, and
  S.~Zafeiriou, ``Agedb: the first manually collected, in-the-wild age
  database,'' in \emph{Proceedings of the IEEE Conference on Computer Vision
  and Pattern Recognition Workshops}, 2017, pp. 51--59.

\bibitem{sengupta2016frontal}
S.~Sengupta, J.-C. Chen, C.~Castillo, V.~M. Patel, R.~Chellappa, and D.~W.
  Jacobs, ``Frontal to profile face verification in the wild,'' in \emph{2016
  IEEE Winter Conference on Applications of Computer Vision (WACV)}.\hskip 1em
  plus 0.5em minus 0.4em\relax IEEE, 2016, pp. 1--9.

\bibitem{Founds2010NIST}
A.~P.~Founds, N.~Orlans, G.~Whiddon, and C.~Watson, ``Nist special database 32
  multiple encounter dataset ii (meds-ii),'' National Institute of Standards
  and Technology, Report NISTIR 7807, 2010.

\bibitem{gross2010multi}
R.~Gross, I.~Matthews, J.~Cohn, T.~Kanade, and S.~Baker, ``Multi-pie,''
  \emph{Image and Vision Computing}, vol.~28, no.~5, pp. 807--813, 2010.

\bibitem{viola2001rapid}
P.~Viola and M.~Jones, ``Rapid object detection using a boosted cascade of
  simple features,'' in \emph{Proceedings of the 2001 IEEE computer society
  conference on computer vision and pattern recognition. CVPR 2001},
  vol.~1.\hskip 1em plus 0.5em minus 0.4em\relax IEEE, 2001, pp. I--I.

\bibitem{girshick2015fast}
R.~Girshick, ``Fast r-cnn,'' in \emph{Proceedings of the IEEE international
  conference on computer vision}, 2015, pp. 1440--1448.

\bibitem{ren2015faster}
S.~Ren, K.~He, R.~Girshick, and J.~Sun, ``Faster r-cnn: Towards real-time
  object detection with region proposal networks,'' in \emph{Advances in neural
  information processing systems}, 2015, pp. 91--99.

\bibitem{zhang2016joint}
K.~Zhang, Z.~Zhang, Z.~Li, and Y.~Qiao, ``Joint face detection and alignment
  using multitask cascaded convolutional networks,'' \emph{IEEE Signal
  Processing Letters}, vol.~23, no.~10, pp. 1499--1503, 2016.

\bibitem{uijlings2013selective}
J.~R. Uijlings, K.~E. Van De~Sande, T.~Gevers, and A.~W. Smeulders, ``Selective
  search for object recognition,'' \emph{International journal of computer
  vision}, vol. 104, no.~2, pp. 154--171, 2013.

\bibitem{zhang2017s3fd}
S.~Zhang, X.~Zhu, Z.~Lei, H.~Shi, X.~Wang, and S.~Z. Li, ``S3fd: Single shot
  scale-invariant face detector,'' in \emph{Proceedings of the IEEE
  International Conference on Computer Vision}, 2017, pp. 192--201.

\bibitem{li2019dsfd}
J.~Li, Y.~Wang, C.~Wang, Y.~Tai, J.~Qian, J.~Yang, C.~Wang, J.~Li, and
  F.~Huang, ``Dsfd: dual shot face detector,'' in \emph{Proceedings of the IEEE
  Conference on Computer Vision and Pattern Recognition}, 2019, pp. 5060--5069.

\bibitem{chen2020yolo}
W.~Chen, H.~Huang, S.~Peng, C.~Zhou, and C.~Zhang, ``Yolo-face: a real-time
  face detector,'' \emph{The Visual Computer}, pp. 1--9, 2020.

\bibitem{turk38}
M.~Turk, A.~Pentland, P.~Belhumeur, and J.~Hespanha, ``Eigenfaces for
  recognition: Journal of cognitive neurosicence,'' 1991.

\bibitem{belhumeur37}
P.~N. Belhumeur, J.~P. Hespanha, and D.~J. Kriegman, ``Eigenfaces vs.
  fisherfaces: Recognition using class specific linear projection,'' \emph{IEEE
  Transactions on pattern analysis and machine intelligence}, vol.~19, no.~7,
  pp. 711--720, 1997.

\bibitem{liu39}
C.~Liu and H.~Wechsler, ``Gabor feature based classification using the enhanced
  fisher linear discriminant model for face recognition,'' \emph{IEEE
  Transactions on Image processing}, vol.~11, no.~4, pp. 467--476, 2002.

\bibitem{Schroff30}
F.~Schroff, D.~Kalenichenko, and J.~Philbin, ``Facenet: A unified embedding for
  face recognition and clustering,'' in \emph{Proceedings of the IEEE
  conference on computer vision and pattern recognition}, Conference
  Proceedings, pp. 815--823.

\bibitem{Liu20}
W.~Liu, Y.~Wen, Z.~Yu, M.~Li, B.~Raj, and L.~Song, ``Sphereface: Deep
  hypersphere embedding for face recognition,'' in \emph{Proceedings of the
  IEEE conference on computer vision and pattern recognition}, Conference
  Proceedings, pp. 212--220.

\bibitem{Wu34}
X.~Wu, R.~He, Z.~Sun, and T.~Tan, ``A light cnn for deep face representation
  with noisy labels,'' \emph{IEEE Transactions on Information Forensics and
  Security}, vol.~13, no.~11, pp. 2884--2896, 2018.

\bibitem{Deng5}
J.~Deng, J.~Guo, N.~Xue, and S.~Zafeiriou, ``Arcface: Additive angular margin
  loss for deep face recognition,'' in \emph{Proceedings of the IEEE Conference
  on Computer Vision and Pattern Recognition}, Conference Proceedings, pp.
  4690--4699.

\bibitem{Gross10}
R.~Gross, I.~Matthews, J.~Cohn, T.~Kanade, and S.~Baker, ``Multi-pie,''
  \emph{Image and Vision Computing}, vol.~28, no.~5, pp. 807--813, 2010.

\bibitem{Wang2018CosFace}
H.~{Wang}, Y.~{Wang}, Z.~{Zhou}, X.~{Ji}, D.~{Gong}, J.~{Zhou}, Z.~{Li}, and
  W.~{Liu}, ``Cosface: Large margin cosine loss for deep face recognition,'' in
  \emph{2018 IEEE/CVF Conference on Computer Vision and Pattern Recognition},
  2018, pp. 5265--5274.

\bibitem{liu2015deep}
Z.~Liu, P.~Luo, X.~Wang, and X.~Tang, ``Deep learning face attributes in the
  wild,'' in \emph{Proceedings of the IEEE international conference on computer
  vision}, 2015, pp. 3730--3738.

\bibitem{daugman1998recognizing}
J.~Daugman, ``Recognizing people by their iris patterns,'' \emph{Information
  Security Technical Report}, vol.~3, no.~1, pp. 33--39, 1998.

\bibitem{maaten2008visualizing}
L.~v.~d. Maaten and G.~Hinton, ``Visualizing data using t-sne,'' \emph{Journal
  of machine learning research}, vol.~9, no. Nov, pp. 2579--2605, 2008.

\end{thebibliography}

\end{document}